\documentclass[10pt,journal,compsoc]{IEEEtran}
\ifCLASSOPTIONcompsoc
  \usepackage[nocompress]{cite}
\else
  \usepackage{cite}
\fi

\ifCLASSINFOpdf
  \usepackage[pdftex]{graphicx}
\else
\fi
\usepackage{amsmath}
\usepackage{dsfont}
\DeclareMathOperator*{\EX}{\mathds{E}}
\allowdisplaybreaks

\usepackage[titlenumbered,ruled]{algorithm2e}
\usepackage{ragged2e}
\usepackage{url}
\usepackage{hyperref}

\usepackage{xcolor}
\makeatletter
\newcommand{\subalign}[1]{%
  \vcenter{%
    \Let@ \restore@math@cr \default@tag
    \baselineskip\fontdimen10 \scriptfont\tw@
    \advance\baselineskip\fontdimen12 \scriptfont\tw@
    \lineskip\thr@@\fontdimen8 \scriptfont\thr@@
    \lineskiplimit\lineskip
    \ialign{\hfil$\m@th\scriptstyle##$&$\m@th\scriptstyle{}##$\hfil\crcr
      #1\crcr
    }%
  }%
}
\makeatother

\newcommand{\norm}[1]{\left\lVert#1\right\rVert}
\usepackage{soul}

\begin{document}
\bstctlcite{IEEEexample:BSTcontrol}
%
\title{Learning Transferable Concepts \\ in Deep Reinforcement Learning}
%
%
%
%

\author{Diego~Gomez,
        Nicanor~Quijano,~\IEEEmembership{Senior~Member,~IEEE,}
        and~Luis~Felipe~Giraldo,~\IEEEmembership{Member,~IEEE,}
\IEEEcompsocitemizethanks{\IEEEcompsocthanksitem D. Gomez, N. Quijano, and L. F. Giraldo are with the Departmento de Ingeniería Eléctrica y Electrónica, Universidad de Los Andes, Bogotá, Colombia.
Corresponding e-mail: df.gomez2a@gmail.com 
}
\thanks{Manuscript submitted to the IEEE Transactions on Pattern Analysis and Machine Intelligence, 23 of february, 2021. Copyright may be transferred without notice, after which this version may no longer be accessible.}
}

\IEEEtitleabstractindextext{%
\justify
\begin{abstract}
While humans and animals learn incrementally during their lifetimes and exploit their experience to solve new tasks, standard deep reinforcement learning methods specialize to solve only one task at a time. As a result, the information they acquire is hardly reusable in new situations. Here, we introduce a new perspective on the problem of leveraging prior knowledge to solve future tasks. We show that learning discrete representations of sensory inputs can provide a high-level abstraction that is common across multiple tasks, thus facilitating the transference of information. In particular, we show that it is possible to learn such representations by self-supervision, following an information theoretic approach. Our method is able to learn concepts in locomotive and optimal control tasks that increase the sample efficiency in both known and unknown tasks, opening a new path to endow artificial agents with generalization abilities.
\end{abstract}

}

\maketitle

\IEEEdisplaynontitleabstractindextext

%
\IEEEpeerreviewmaketitle

\IEEEraisesectionheading{\section{Introduction}\label{sec:introduction}}

%
%
%
%

 

\IEEEPARstart{H}{umans} and animals alike have evolved to complete a wide variety of tasks in a world where resources are scarce. In particular, since learning and planning have an associated energy cost, the brain has probably evolved to solve multiple tasks while spending the least possible amount of energy \cite{mery2004operative, niven2016neuronal}. It is only reasonable then that the brain has an innate ability to generalize what it learns in one task to succeed in future ones. Otherwise, it would be too costly to learn from scratch the appropriate solution for each problem encountered \cite{bongard2006resilient, haarnoja2018soft}. Given how any artificial agent would face exactly the same burdens, it is highly desirable for it to possess the same generalization capacities. Deep reinforcement learning techniques have shown outstanding progress in solving complex tasks \cite{mnih2015human, silver2017mastering} and there has been advances in applying these methods to produce real life adaptation \cite{bongard2006resilient, haarnoja2018soft, andrychowicz2020learning}, even under significant body alterations \cite{cully2015robots}. However, there is still much progress to do in terms of transferring knowledge among multiple tasks \cite{flesch2018comparing, cobbe2019quantifying, zhao2019investigating} and under constraints like time \cite{harb2018waiting}, memory capacity, and energy.  


One of the common traits of standard deep reinforcement learning methods is that sensory inputs and their successive encodings are represented as continuous real-valued vectors. This type of distributed representations is very flexible and allows using efficient gradient-based optimization techniques, both properties that greatly enhance the learning performance in single tasks. However, this flexibility encourages learning very complex models, which usually take advantage of spurious statistical patterns \cite{duan2016benchmarking, geirhos2020shortcut} that are not essential to solve the task and that are not present in similar tasks, or even in the same task when sensory inputs are noisy. Thus, the excess in flexibility directly inhibits the transference of knowledge between tasks. In contrast, both animals and humans exhibit the use of disentangled and discrete representations, concepts, to encode sensory inputs and internal states \cite{zhang2008discrete, linderman2019hierarchical, higgins2020unsupervised}. These concepts, on the contrary, favor low-complexity models that capture only the most relevant patterns \cite{zaslavsky2018efficient}. For example, the use of a discrete set of colors to identify edible food \cite{briscoe2001evolution} is a useful trait for survival that separates sensory inputs into discrete categories while ignoring essentially useless information. Aside from promoting simplicity, concepts tend to be modular and transferable between different contexts. The human language is perhaps the epitome of such traits: humans seem to possess an innate ability to represent arbitrarily complex phenomena from a finite number of discrete expressions \cite{hauser2002faculty}. Discovering methods that identify rich discrete representations, like colors, language expressions, or objects, seems then to be a promising path to endow artificial agents with the ability of generalization \cite{shanahan2020artificial, davidson2020investigating}.  


Past artificial intelligence techniques relied on the use of discrete symbolic representations that were the result of hand-picked features or expert-based rules. These techniques were later demonstrated to be sub-optimal in comparison with fully learning-based methods that made no assumptions about the problem being solved \cite{silver2017mastering, tesauro2002programming}. Correspondingly, current attempts to reintroduce the use of discrete representations let the models learn by themselves such representations. Instead of directly providing them, an inductive bias, in the form of an architecture decision or an optimization criterion, is introduced in the algorithm to guide the representation learning \cite{hessel2019inductive}. There are several examples, such as the inductive bias of objectness \cite{veerapaneni2019entity, kulkarni2019unsupervised}, the use of finite vector codes with continuous entries \cite{razavi2019generating}, the hierarchical break down of problems through localized policies, commonly called options or skills \cite{sutton1999between, bacon2017opctioncritic, frans2018meta, goyal2020reinforcement}, and the assumption of the existence of discrete world models \cite{hafner2020mastering}. In this work we take a similar approach. We propose an information theoretic loss function to learn discrete representations off-policy, in a self-supervised manner. We prove that this loss corresponds to an upper bound of the regret and empirically show that the associated representations make more efficient the process of learning policies in the same tasks where the representations are learned and in future unseen tasks.         


Our main contributions are threefold: first, to the best of our knowledge, this is the first work that proves theoretically that the mutual information between states and actions, conditioned on learned representations and contexts, is an optimal upper bound for how suboptimal is a policy that ignores the states and only relies on their representation to select actions. Second, we derive the conditions in which the abstractions learned by minimizing this mutual information bound
are expected to be transferable. We show empirically that using the representations increases the sample efficiency of the policy learning process, even when using different dynamics and drastically different reward functions. And third, we show that the learned representations correspond to intuitive concepts related with objects and relative positions, that they can present compositionality, and that a similar loss function that ignores the information of the context precludes learning this type of concepts.

\section{Concept learning}
\subsection{Multi-task and transfer learning}
Before further elaborating on the problem of finding useful discrete representations, a brief description of the multi-task and transfer problems in reinforcement learning is needed. In the multi-task problem, an agent finds itself in an environment and their joint state is specified by a vector $s\in \mathcal{S}$ (where $\mathcal{S}$ denotes the set of states), which corresponds to any input information required to understand the evolution of the system. The agent has the ability to change this state to a new state $s'\in \mathcal{S}$ by taking an action $a$ from a set of actions $\mathcal{A}$ (for example, through the activation of actuators or the selection between different decisions), and, as a result of this change, the agent receives a reward $r\in\mathds{R}$. During its lifetime, the agent faces a sequence of tasks from a set $\mathcal{C}$, where each one determines how the states will evolve and what rewards will the agent receive (for example, decision or control problems, or games). So, the agent has to find a policy, a behavior rule that selects actions based on states, for each task $c\in\mathcal{C}$, such that the accumulated reward during its whole lifetime is maximized. This can be put more succinctly by stating that the agent tries to solve a control problem in a contextual Markov Decision Process (CMDP) \cite{kirk2021survey}. This means that we have a tuple $\mathcal{M}=(\mathcal{S},\mathcal{A},\mathcal{C},r,T,p_{C},p_{S_0},\gamma)$, where $\mathcal{S}, \mathcal{A},$ and $\mathcal{C}$ denote the sets of states, actions, and tasks (or contexts), $r:\mathcal{S}\times\mathcal{A}\times\mathcal{C}\to\mathds{R}$ is the reward function, $T:\mathcal{S}\times\mathcal{A}\times\mathcal{C}\to\Delta(\mathcal{S})$ is the transition map, $p_{C}\in\Delta(\mathcal{C})$ the distribution of tasks, $p_{S_0}:\mathcal{C}\to\Delta(\mathcal{S})$ is the initial distribution of states, and $\gamma$ is a discount factor. The objective of the agent is to find the optimal policy $\pi^*:\mathcal{S}\times\mathcal{C}\to\Delta(\mathcal{A})\in\Pi(\mathcal{M})$ that maximizes the expected cumulative reward
$ R_{\mathcal{M}}(\pi) := \mathop{\EX}_{C\sim p_C}\sum_{t=0}^\infty \gamma^t\mathop{\EX}_{\subalign{\vspace{0.001cm}\\S_0&\sim p_{S_0}\\S_{t+1}&\sim T\\A_t&\sim\pi}}\left[r(s_t,a_t,c)\right]\text{ .}$
To do so, the agent uses a learning algorithm $L:\text{M}\to\Pi$ that specifies how to interact with $\mathcal{M}\in\text{M}$ to obtain a policy $\pi\in\Pi(\mathcal{M})$ with small regret, defined as $\text{regret}_{\mathcal{M}}(\pi)=R_{\mathcal{M}}(\pi^*)-R_{\mathcal{M}}(\pi)$.
How good is a learning algorithm $L$ depends on the regret achieved, but also of its sample efficiency, stability and related properties. We will denote the performance of $L$ in a particular problem $\mathcal{M}$ as $J_{\mathcal{M}}(L)$.

The transfer learning problem concerns itself with finding learning algorithms $L_{\text{transfer}}:\text{M}\times\text{M}\to\Pi$ that are able to use information from a training CMDP $\mathcal{M}_{\text{train}}$ to improve its performance $J_{\mathcal{M}_{\text{test}}}\left(L_{\text{transfer}}\right)$ in a test CMDP $\mathcal{M}_{\text{test}}$, with respect with some baseline $L_{\text{baseline}}:\text{M}\to\Pi$ that only uses information from $\mathcal{M}_{\text{test}}$ \cite{kirk2021survey, taylor2009transfer, zhu2020transfer}.

\begin{figure}
\centering
\includegraphics[width=\columnwidth]{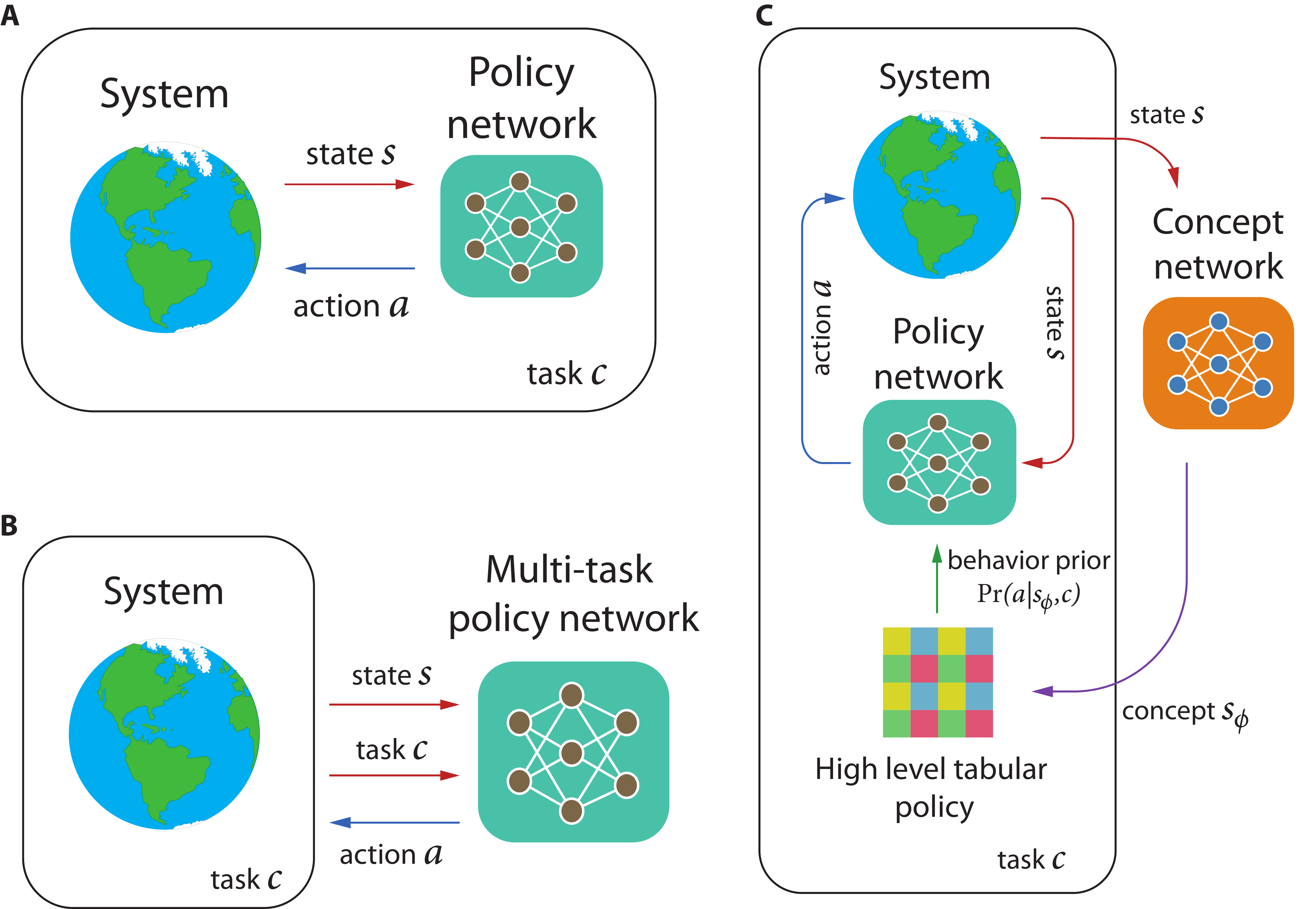}
\caption{Deep reinforcement learning methods in the context of multi-task learning. (A) Standard methods learn, for each task $c$, the parameters of a prototypical neural network that determines the policy $\pi(a|s,c)$. (B) Goal-based methods parametrize the task as a goal vector that allows using a single neural network to perform multiple tasks. This usually makes learning harder because of interference between tasks. (C) Concept-based methods learn task-agnostic parameters related to the actions and to a classifier of states (the set of categories $S_\phi$ of this classifier are the concepts). A task-dependent tabular policy $\pi_\phi(a|s_\phi,c)$ may provide a prior to learn faster the policies $\pi(a|s,c)$.}
\label{fig:methods}
\end{figure}

Multiple approaches have been proposed to address the multi-task and transfer learning problems \cite{zhang2019deep}. The naive approach in the multi-task case would be to find a policy for each possible task $c$, which in and of itself is a challenging problem \cite{mnih2015human, haarnoja2018soft}. However, these learning algorithms rely on single-use models that cannot be exploited even in very similar tasks (Fig. \ref{fig:methods}A). In the case that the context can be represented as a vector and it is accessible to the agent, the opposite approach consists in learning a single policy that works in all tasks $\mathcal{C}$. While in principle this is the desired solution, it is generally more difficult to train these models due to interference between optimization objectives and ultimately the approach might not result in an improvement in neither efficiency nor performance \cite{yu2020gradient} (Fig. \ref{fig:methods}B).

\textit{Meta-learning} is an intermediate approach were the initial test parameters of the models are learned in $\mathcal{M}_{\text{train}}$ in such a way that they optimize the learning efficiency in $\mathcal{M}_{\text{test}}$ \cite{finn2017model, nichol2018first, frans2018meta}. Alternatively, \textit{imitation learning} proposes to use data from experts to infer faster the appropriate behavior, taking into account that the $\mathcal{M}_{\text{train}}$ where the examples were generated might differ from $\mathcal{M}_{\text{test}}$ \cite{kostrikov2019imitation}. A third family of methods involves \textit{learning skills}, which are localized policies $\left.\pi\right|_{\mathcal{S}'}:\mathcal{S}'\times\mathcal{C}\to\Delta(\left.\mathcal{A}\right|_{\mathcal{S}'})$ restricted to $\mathcal{S}'\subset\mathcal{S}$ and that ideally do not depend on the context $c$ \cite{riemer2018learning, eysenbach2019diversity}. The goal is to separate the state-space in small regions where is easier to learn local skills and to re-use them in different contexts. An additional approach consists in \textit{learning explicit state abstraction maps} $\phi:\mathcal{S}\to\Delta(\mathcal{S}_\phi)$, where $\mathcal{S}_\phi$ is ideally a smaller space than the original $\mathcal{S}$. Similarly to the skill learning case, the objective is to reduce the learning complexity by separating the original policy in the abstraction map $\phi$ and an abstract policy $\pi_\phi:\mathcal{S}_\phi\times\mathcal{C}\to\Delta(\mathcal{A})$ \cite{singh1995reinforcement, li2006towards}.  




\subsection{Concepts as equivariant sets of behavior}\label{sec:equiv}


As detailed in the \hyperref[sec:introduction]{Introduction}, rich discrete representations might be a key factor to handle the problem of transfer learning. Thus, in this work we propose an algorithm to learn a state abstraction map $\phi$, that maps states $s\in\mathcal{S}$ to concepts $s_\phi\in\mathcal{S}_\phi$, where $|\mathcal{S}_\phi|=N_S$. We refer to this problem as concept learning since the objective is to learn composable discrete representations of a highly dimensional continuous state-space $\mathcal{S}$ that are transferable between different contexts (Fig. \ref{fig:methods}C). Our theoretical results will limit to a finite state-space for simplicity, but our experiments will be carried out in vector and pixel spaces.

To motivate the idea of concepts, let us consider the following example. Suppose that you want to make a round-trip to city $c_1$ and that you rent a car $s$ for this purpose. You can take routes $a_1$ or $a_2$. Taking route $a_1$ results in a shorter trip and there is available a fuel station, while in $a_2$ an electric charging point. In both cases the trip to city $c_1$ takes only a few minutes and so there is no need to stop. One could argue route $a_1$ is better, no matter which car $s$ was given to you by the rental company, since it is shorter. Now, suppose that you want to make a round-trip to a second city $c_2$. This city is farther than $c_1$, but you can reach it by taking routes $a_1$ or $a_2$ as well. In this case, the energy will not be enough for the whole trip and you will have to stop to get more fuel or charge the vehicle. In this second case the particular car provided by the rental company, or, specifically, if it is an electric or a combustion engine vehicle, is relevant to determine the optimal action. The reverse is also true: a route being optimal for a specific car $s$ means that a second car $s'$ is different to $s$ if the optimal route is a different one. Now imagine that you have no notion of a car being electric or having a combustion engine, and that you have no access to information about what makes different the routes $a_1$ and $a_2$. In this case, you would have to explore both routes in multiple trips, to make an estimate of which route is optimal, and you would have to do this for each possible car $s$. This means that the number of required trips to learn the optimal strategy would be proportional to the number of cars that you could rent. This would be the same for any other city $c$ that you would like to consider. However, there is a simpler way to find the optimal strategy for all possible cars: if you are able to correctly classify them in an electric class $s_{\phi,1}$ or a combustion engine class $s_{\phi,2}$, then you would be able to learn what to do for all electric cars corresponding to $s_{\phi,1}$ having estimated the optimal action for only a single electric car, and similarly for the combustion engine cars. So, the required number of trips to learn the best strategy would be proportional not to the entire number of cars, but to the number of classes of cars, which in this case is a much more reduced number. Now, how can you learn to distinguish between electric and combustion engine cars? If we relied only on the information of the trip $c_1$, knowing the optimal policy would not be useful, since the action to take in each car would be the same. However, if we relied in the information of task $c_2$, we would be able to determine which car is electric if its corresponding optimal action is taking route $a_2$. In this way, learning the concept of electric / combustion engine vehicle would allow to learn optimal routes much more efficiently in new cities.


\begin{figure}
\centering
\includegraphics[width=\columnwidth]{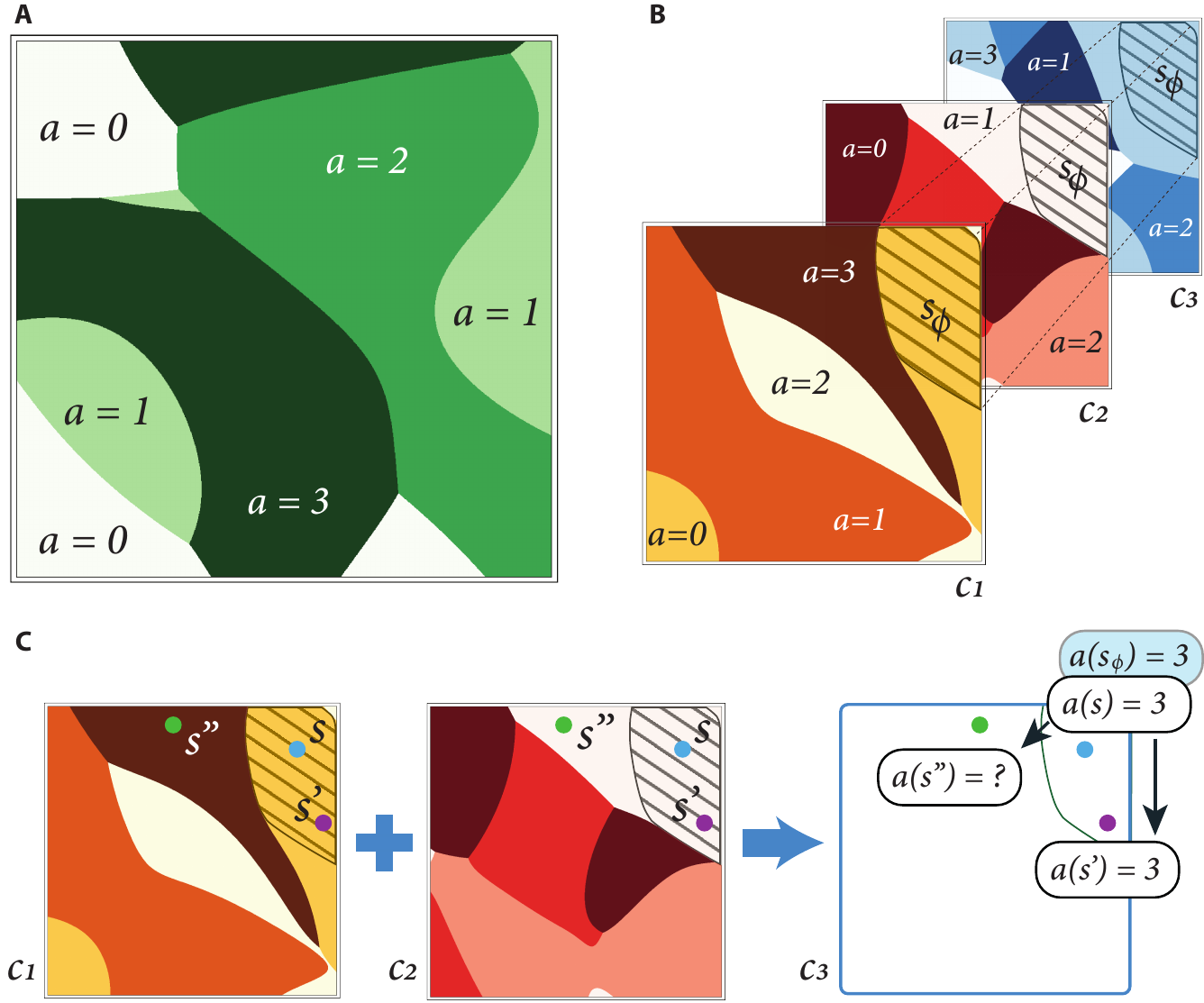}
\caption{Concepts as equivariant sets. (A) A deterministic policy from states to actions $a(s,c)$ determines a partition of the state-space in multiple regions $s_\phi$ when a single task is considered. States corresponding to the same action (\textit{i.e.}, in the same region) can be considered equivalent (in the case of a non-deterministic policy $p(a|s,c)$, the states would belong to each partition $s_\phi\in \mathcal{S}_\phi$ with certain probability). (B) In a multitask setting, the concepts correspond to the intersections of the partitions provided by the policies. States corresponding to the same intersection $s_\phi$ can be considered equivalent since the agent behaves the same in them, only depending on the task. (C) For example, $s$, $s'$, and $s''$ correspond to the same action in the task $c_2$, but in $c_1$ the action in $s''$ is different. Consequently, in a third task, information about the action of $s$ would provide us with more information about the action to take in $s'$ than the action in $s''$.}
\label{fig:concepts}
\end{figure}


Put in more general terms, consider an agent in a transfer learning setting that can train in a CMDP $\mathcal{M}_\text{train}$ and will be tested in $\mathcal{M}_\text{test}$. The agent has the ability to classify states $s$ into concepts $s_\phi$, by means of a state-abstraction classifier $\phi$. We want the agent to learn a classifier $\phi$ such that its selection of an action $a$ in the future task depends as little as possible on the state $s$ once the concept $s_\phi$ is known (just as specific features of a car became irrelevant once one knew it was electric or not). That is, we want the concepts to provide most of the necessary information to decide when to use each action. As we saw in our car example, this problem can be solved in a single task by looking at how the optimal policy in a task partitions the state-space: if two states $s$ and $s'$ correspond to the same optimal action $a(s)$, then they can be seen as equivalent and therefore assigned to the same concept $s_\phi$ (Fig. \ref{fig:concepts}A). However, in a multitask setting, this does not hold since two states corresponding to the same action in a specific task may not correspond to the same action in a second task (just as the combustion engine cars corresponded to the same action that electric cars did in task $c_1$ but not in task $c_2$). In this case, it is the intersection of the partitions generated by the optimal policies the sets of states that can be understood as concepts (Figs. \ref{fig:concepts}B, C). This is the case since the behavior of any two states $s$ and $s'$ corresponding to the same concept $s_\phi$ will be the same in each task $c$, \textit{i.e.}, $a(s,c)=a(s',c)$. Something to note here is that the action that is common between the two states does depend on the task $c$ and for this reason the concepts are task-\textit{equivariant} sets of behavior, as opposed to just \textit{invariant} sets where the same action is always chosen without considering the task information.

Up to this point, the way in which we propose to classify states as concepts can be understood as a multi-task extension of the $\pi^*$-irrelevance abstraction $\phi_{\pi^*}$ introduced in \cite{jong2005state}. However, just as what happens with similar state abstractions, the requirement that $a^*(s,c)=a^*(s',c)$ for any context $c$ is not scalable to continuous domains. To make tractable the problem, we will use a probabilistic framework where two states $s$ and $s'$ will be similar if the mean divergence of $\pi(a|s,c)$ and $\pi(a|s',c)$ is small, instead of requiring that $a^*(s,c)=a^*(s',c)$. In this sense, our method is a multi-task probabilistic relaxation of the $\pi^*$-irrelevance abstraction, similar to how other methods are relaxations of $Q^*-$irrelevance abstractions or bisimulation equivalence classes \cite{abel2016near, zhang2021learning}. Aside from the natural interpretation of concepts as task-equivariant sets of behavior, we consider that other state abstractions, such as the $Q^*-$irrelevance, enforce constraints that are not natural and as a result are less transferable than those representations derived from the $\pi^*$-irrelevance notion, which is one of the less restrictive \cite{li2006towards}. For example, by asking for similar Q-values one necessarily is assuming that reward functions are similar between tasks.

Moreover, given that the number of tasks $|\mathcal{C}|$ can be potentially uncountable (and so the number of potential concepts $N_S$), in practice the existence of a useful state-abstraction classifier $\phi$ relies on the assumption that the tasks of interest lie in a very particular subset of all the task-space, where there exist large clusters of states such that their interaction with actions is similar across tasks. This is not unlike humans, which seem to present innate priors related to the existence and persistence of physical objects \cite{baillargeon2008innate, fields2017eigenforms}, as well as the already mentioned abilities of color perception and language abstraction. 

As a solution to the concept learning problem we pose, we take two steps: first, to train the agent in the available set of tasks following standard methods; and second, to use the policies learned to generate triples of examples $(c,s,a)$, which will be used to train the classifier $\phi$, following an information theoretic approach, as explained below.

\subsection{Learning representations with mutual information}
A random variable $X$ can take several values, according to its corresponding distribution $p(X)$. Depending on this distribution, we may have more or less uncertainty about which specific value it can take. A measure for this uncertainty is the entropy $H(X)=\mathds{E}_{p(X)}\left[-\log p(x)\right]$ (where $\mathds{E}\left[\cdotp\right]$ denotes the expected value of the function between brackets), which is maximal when the distribution is uniform (for a bounded variable) and minimal when the probability of a specific value is 1. Now, let us consider a second random variable $Y$. In general, these two variables might be correlated, and if we get to know the value that $Y$ takes, we may exploit this knowledge to reduce our uncertainty of what value could $X$ take. This uncertainty can be measured with the conditional entropy $H(X|Y)=\mathds{E}_{p(X,Y)}\left[-\log p(x|y)\right]$. The mutual information between the two random variables is precisely the difference between these two entropies:
\begin{equation*}
    \resizebox{0.82\columnwidth}{!}{$
    I(X:Y) = 
    H(X) - H(X|Y) = 
    \mathds{E}_{p(X,Y)}\left[\log \frac{p(x,y)}{p(x)p(y)}\right]\text{ .}$}
\end{equation*}
Thus, the mutual information $I(X:Y)$ captures how much information is acquired about one variable when we know the value of the other one. Additionally, one might consider the mutual information between two random variables $X$ and $Y$, given the knowledge of a third random variable $Z$. This is the conditional mutual information: $I(X:Y|Z)=H(X|Z)-H(X|Y,Z)$.

In machine learning, one typically desires to learn representations of inputs that are useful in downstream tasks while balancing complexity and performance \cite{kingma2014auto, razavi2019generating}. The concept of mutual information provides an optimality measure to learn representations, since maximizing it results in an encoding of inputs of minimum length, maximal disentanglement, or maximal prediction power \cite{linsker1988self, schmidhuber1992learning, tishby2000information, alemi2017deep, hafner2020action}. For this reason, recent works in deep reinforcement learning make use of mutual information optimization criterions to facilitate learning optimal policies \cite{goyal2019infobot, igl2019generalization, goyal2020reinforcement}. We do the same here, taking into account our goal of learning concepts that are, as much as possible, equivariant sets of behavior. In concrete, we propose to learn the classifier of states $\phi^*$ that solves:
\begin{equation}\label{eq:proposed_metric}
    \resizebox{0.45\columnwidth}{!}{$
    \phi^*\in\min_{\phi}I_{\mathcal{M}}(S:A|S_\phi,C)\hspace{2pt}\text{ ,}$}
\end{equation} where $S,S_\phi,A,$ and $C$ are random variables corresponding to the state $s$, concept $s_\phi$, action $a$, and context $c$, respectively\footnote{Here, the distribution $p_S(s|c)$ corresponds to the expected discounted state distribution $p_S(s|c):= (1-\gamma)\sum_{t=0}^\infty\gamma^t\text{Pr}(s_t=s|\pi^*,c)$ that results from following the optimal policy $\pi^*(a|s,c)$ under the dynamics $T(s_{t+1}|s_t,a_t,c)$, and the initial state distribution $p_{S_0}(s_0|c)$.}. In this manner, we want the classifier that makes the state $s$ the least informative as possible about the action $a$ once the concept $s_\phi$ and the context $c$ are known\footnote{The objective in Eq. (\ref{eq:proposed_metric}) is very similar to the information bottleneck \cite{tishby2000information} applied to the Markov chain $S_\phi\leftarrow S\rightarrow A$, but conditioned on the task $C$. We develop this connection on the Supplementary Material.}. While the motivation of our bound came from the idea of concepts as task-equivariant sets that try to tackle the transfer learning problem, this loss function can be related to the problems of imitation learning and the trade-off between compression and performance \cite{tishby2011information, rubin2012trading, abel2019state, asadi2020learning}.  




\subsection{Learning concepts through information minimization is optimal}
While intuitive, the optimization problem we propose in Eq. (\ref{eq:proposed_metric}) might seem arbitrary. In this section, we will provide a basis from a control standpoint. 
First, in Lemma 1 we relate the statistical distance between the optimal policy $\pi^*$ and an arbitrary policy $\pi_\phi$ to their difference in performance.


\textbf{Lemma 1:}\label{lem1} \textit{Let the tuple $\mathcal{M}=(\mathcal{S},\mathcal{A},\mathcal{C},r,T,p_{C},p_{S_0},\gamma)\in M$ be a CMDP with finite $\mathcal{S},\mathcal{A},\mathcal{C}$, $\mathcal{S}_\phi$ a finite set of concepts, $\phi(s)= p_{S_\phi}(\cdotp|s)$ be the probability distribution of concepts given the state $s\in S$, and $\pi_\phi(s,s_\phi,c)=p_A(\cdotp|s,s_\phi,c)$ the agent's policy. Then, using a policy $\pi_\phi$, instead of an optimal policy $\pi^*\in\arg\max_{\pi}R_\mathcal{M}(\pi)$, results in a regret that is bounded by the expected Kullback-Leibler divergence with $\pi^*$. That is, $$\text{regret}_{\mathcal{M}}(\pi)^2\leq F_{\mathcal{M}}\cdotp \mathop{\EX}_{C,S,S_\phi}\left[\mathcal{D}_{KL}(\pi^*||\pi_\phi)\right],$$ where $F_{\mathcal{M}}=2\norm{r}_\infty/(1-\gamma)^2$, and $\mathcal{D}_{KL}(p||q)=\sum_ip_i\log\frac{p_i}{q_i}$ is the Kullback-Leibler divergence of the distributions $p$ and $q$.}

Lemma 1 says that the closest a policy is from the optimal policy, the better we can expect it to perform. So, if one does know the optimal policy but for some reason wants to restrict the information used to take decisions, the best one can do is trying to make the restricted policy as close as possible to the optimal policy. The following theorem shows that the best policy that only considers concepts, instead of states, is the one that corresponds to the objective in Eq. (\ref{eq:proposed_metric}).


\textbf{Theorem 1:}\label{th1} \textit{Let $\mathcal{M}$, $\mathcal{S}_\phi$, $\phi$, and $\pi_\phi$ be as in \hyperref[lem1]{\textnormal{\textbf{Lemma 1}}}, and let us define $U_{\pi^*}(\pi)=F_{\mathcal{M}}\cdotp \mathop{\EX}_{C,S,S_\phi}\left[\mathcal{D}_{KL}(\pi^*||\pi)\right]$ as the upper bound of the regret in \hyperref[lem1]{\textnormal{\textbf{Lemma 1}}}. If $\pi_\phi$ is constrained to be a function of only the concept $s_\phi$ and the task $c$, i.e., $\pi_\phi(s,s_\phi,c)=\pi_\phi(s_\phi,c)$, then the optimal constrained policy $\pi_\phi^{*}$ is the marginal action distribution 
\begin{equation}\label{eq:marginal_policy} 
    \pi_\phi^{*}(a|s_\phi,c)=\sum_s\pi^*(a|s,s_\phi,c)p_{S}(s|s_\phi,c)\hspace{2pt}\text{ .}
\end{equation}
Correspondingly, the lowest upper bound for the regret is proportional to the mutual information between the state $s$ and the action $a$, conditioned on the concept $s_\phi$ and the task $c$. That is, 
\begin{equation}\label{eq:info_bound}
    \resizebox{0.91\columnwidth}{!}{$
    \text{regret}_{\mathcal{M}}(\pi_\phi^*)^2\leq F_{\mathcal{M}}\cdotp I_{\mathcal{M}}(A:S|S_\phi,C)\leq U_{\pi^*}(\pi_\phi)\hspace{1pt}\text{ , }\forall\pi_\phi\hspace{2pt}\text{ .}$}
\end{equation}}

Inequality (\ref{eq:info_bound}) in \hyperref[th1]{Theorem 1} states that the best possible upper bound on the regret, given a classifier $\phi$, is precisely the mutual information of the optimization problem in Eq. (\ref{eq:proposed_metric}). So, we would like to find the classifier that minimizes this lowest upper bound (see Supplementary Material for an additional comment on how to optimize this quantity). We conclude with the next lemma and corollary by relating the train and test performances.

\textbf{Theorem 2:}\label{th2} \textit{
Let $\mathcal{M}$ and $\phi$ be as in \hyperref[th1]{\textnormal{\textbf{Theorem 1}}}. Also, let us define $D(s,s'|c)=\mathcal{D}_{KL}(\pi^*(s,c)||\pi*(s',c))$, the behavior dissimilarity of a pair of states $(s,s')\in\mathcal{S}^2$, given the context $c\in\mathcal{C}$; and $\mathcal{J}_{\phi}(s,s'|c)=\sum_{s_\phi}\frac{\phi(s_\phi|s)\phi(s_\phi|s')}{p_{S_\phi}(s_\phi|c)}$, the coupling of $s$ and $s'$ in the context $c$ under $\phi$. Then the mutual information associated to $\mathcal{M}$ is bounded as follows:
\begin{equation}\label{eq:new_bound} 
    \resizebox{0.91\columnwidth}{!}{$
    I_{\mathcal{M}}(A:S|S_\phi,C) \leq 
    \mathop{\EX}_{\subalign{\vspace{0.001cm}\\C&\sim p_{C}\\S,S'&\sim p_S}}\left[\mathcal{J}_{\phi}(s,s'|c)D(s,s'|c)\right] \hspace{2pt}\text{ .}$}
\end{equation}
}

\hyperref[th2]{Theorem 2} states that the ideal concept classifier $\phi$ makes the coupling $\mathcal{J}_\phi$ of each pair of states small if they are dissimilar in terms of their associated optimal behavior (this giving priority according to the state visitation). This result is interesting in and of itself since it introduces an upper bound over the regret that probably is simpler to optimize. However, we focus on its implications for transferability.

\textbf{Corollary 1:}\label{cor1} \textit{
Let $\mathcal{M}_{\text{train}}, \mathcal{M}_{\text{test}}\in M$ be two CMDPs that share the state, action, and context sets $(\mathcal{S},\mathcal{A},\mathcal{C})$, and let $\phi_{\text{train}}^*(s)=p_{S_\phi}(\cdotp|s)\in\{0,1\}^{|\mathcal{S}_\phi|}$ be the optimal deterministic probability distribution of concepts in $\mathcal{M}_{\text{train}}$. Then the regret in $\mathcal{M}_{\text{test}}$ is bounded as follows:
\begin{equation}\label{eq:transfer_bound} 
    \frac{1}{F_{\mathcal{M}_\text{test}}}\text{regret}_{\mathcal{M}_\text{test}}(\pi_\phi^*)^2\leq D(s_{\max},s'_{\max}|c_{\max}) \hspace{2pt}\text{ ,}
\end{equation}
where $s_{\max}, s'_{\max}, c_{\max}\in\arg\max_{\mathcal{J}_{\phi_{\text{train}}^*}(s,s',c)>0} D(s,s'|c)$ are the pair of states and context for which the coupling is positive and that maximize the behavior dissimilarity. 
}

The upper bound in inequality (\ref{eq:transfer_bound}) provides an intuition of when are concepts transferable between contexts. Specifically, we only have a guarantee that we can improve the performance in a test task if any two states $s,s'$ that are behaviorally similar in the training set of tasks are also similar in the test set (otherwise the bound is large). The vehicle example of Section \ref{sec:equiv} precisely shows how the first task was insufficient to learn the necessary electric /combustion engine vehicle concept. Conversely, if all the necessary information to distinguish between states is in the training set of tasks, the difference of state visitations becomes of secondary importance, as we show empirically.

\subsection{Transferring concepts}\label{sec:concepts}
In principle, once concepts are learned, one can simply replace states by concepts and use any available reinforcement learning algorithm to learn the abstract policy $\pi_\phi$. In practice, it is not so simple since concepts correspond to observations in a partially observable MDP and consequently bootstrapping methods like Q-learning may not work well.

To address this problem, we propose a Monte Carlo variation. For a given task $c$, the standard algorithm \cite{sutton2018reinforcement} keeps an estimate of the matrix of Q-values $Q_\phi(s_\phi,a)=\mathds{E}_{\pi_\phi(a|s_\phi)}[\sum_{t}^\infty\gamma^tr_t|s_{\phi,t}=s_\phi,a_t=a]$ and a corresponding $\epsilon-$soft policy, which are both updated after sampling a trajectory. The update of the policy consists in assigning to a concept the greedy action $a^*(s_\phi)=\arg\max_aQ_\phi(s_\phi,a)$. However, in our experiments we observed that the policy tended to change abruptly with this update and as a result the estimates of the Q-values diverged.

To prevent the divergence problem, we initialize the policy as a softmax distribution $\pi_{\phi}(a|s_\phi)\propto \exp\left(Q_\phi(s_\phi,a)/\alpha(s_\phi)\right)$, where $\alpha$ is a temperature parameter that regulates the entropy of the policy. In subsequent iterations, we use the same softmax distribution as a target, but we restrict our policy to be close to the previous one, following the trust region idea of prior works \cite{schulman2015trust}:
\begin{align*}
    \pi_{\phi}(\cdotp|s_\phi)\leftarrow\arg\min_p&\mathcal{D}_{KL}\left(p\Big|\Big|\frac{\exp(Q_\phi(s_\phi,\cdotp)/\alpha(s_\phi))}{Z(s_\phi)}\right)\\[0.2cm]
    \text{ such that }&\mathcal{D}_{KL}\left(p||\pi_{\phi}(\cdotp|s_\phi)\right)\leq\epsilon_{\text{MC}}\text{ ,}
\end{align*}
where $\epsilon_{\text{MC}}$ is a threshold hyperparameter. The resultant procedure is summarized in \hyperref[alg1]{Algorithm 1}.

\begin{algorithm}
\caption{Trust Region Monte Carlo (TRMC)}
\label{alg1}
\footnotesize
Initialize matrices $Q_\phi$, $\pi_\phi$, and the initial entropy $H$

\For{iteration$ = 1,2,\cdots$}{
Sample trajectory $\tau=(s_{\phi,0},a_{0},r_{1},\cdots)$ 

Update $Q_\phi$ with every-visit return estimates of $\tau$

\If{should update policy this iteration}{
    Perform line search of $\alpha(s_\phi)$ to ensure $H$
    
    Store target: $\pi_{\text{target}}\leftarrow\frac{\exp(Q_\phi(s_\phi,\cdotp)/\alpha(s_\phi))}{Z(s_\phi)}$
    
    Perform line search for $p$ that minimizes $D_{KL}(p||\pi_{\text{target}})$ while $D_{KL}(p||\pi_{\phi})<\epsilon_{MC}$
    
    Update policy: $\pi_\phi\leftarrow p$
    
    Decrease $H$ as desired
}
}
\end{algorithm}

Our approach will effectively accelerate learning in the case that the concepts allow to estimate $Q_\phi(s_\phi,a)$ faster than standard methods are able to estimate the state-action Q-values. While this is not guaranteed, choosing the classifier that optimizes Eq. (\ref{eq:proposed_metric}) is in a certain sense the best option, according to \hyperref[th1]{Theorem 1} . In addition, we could use the high-level policy $\pi_\phi$ as a behavior prior that informs the learning of policy $\pi$. This can be done by penalizing the difference between both sides of Eq. (\ref{eq:marginal_policy}).

\subsection{Compositional concepts}
So far, the concepts considered determine a partition of the state-space. This is equivalent to say that they are exclusive, \textit{i.e.}, that only one concept can be perceived at a time. Such a supposition imposes learning complex concepts which might not be transferable to new tasks. We propose a simple approach to allow the agent to learn simpler concepts which combined can make up composite concepts. The approach consists in separating the original random variable $S_\phi$ in a tuple of $N_U$ random variables: $S_\phi=(U_0,\cdots,U_{N_U-1})$. Each random variable $U_i$ corresponds to a family of simple concepts and it can be learned by means of a separate classifier $\phi_i$ while trying to optimize Eq. (\ref{eq:proposed_metric}).  

\section{Experiments} 
We conducted multiple experiments to test our proposed method to learn useful discrete representations. In particular, we assess the qualitative characteristics of the concepts, such as their disentanglement or compositionality, their transfer learning performance to the same tasks where the concepts where learned, i.e.,  $\mathcal{M}_{\text{test}}=\mathcal{M}_{\text{train}}$, and to unknown tasks with different reward functions and dynamics.

\begin{figure*}
\centering
\includegraphics[width=0.76\linewidth]{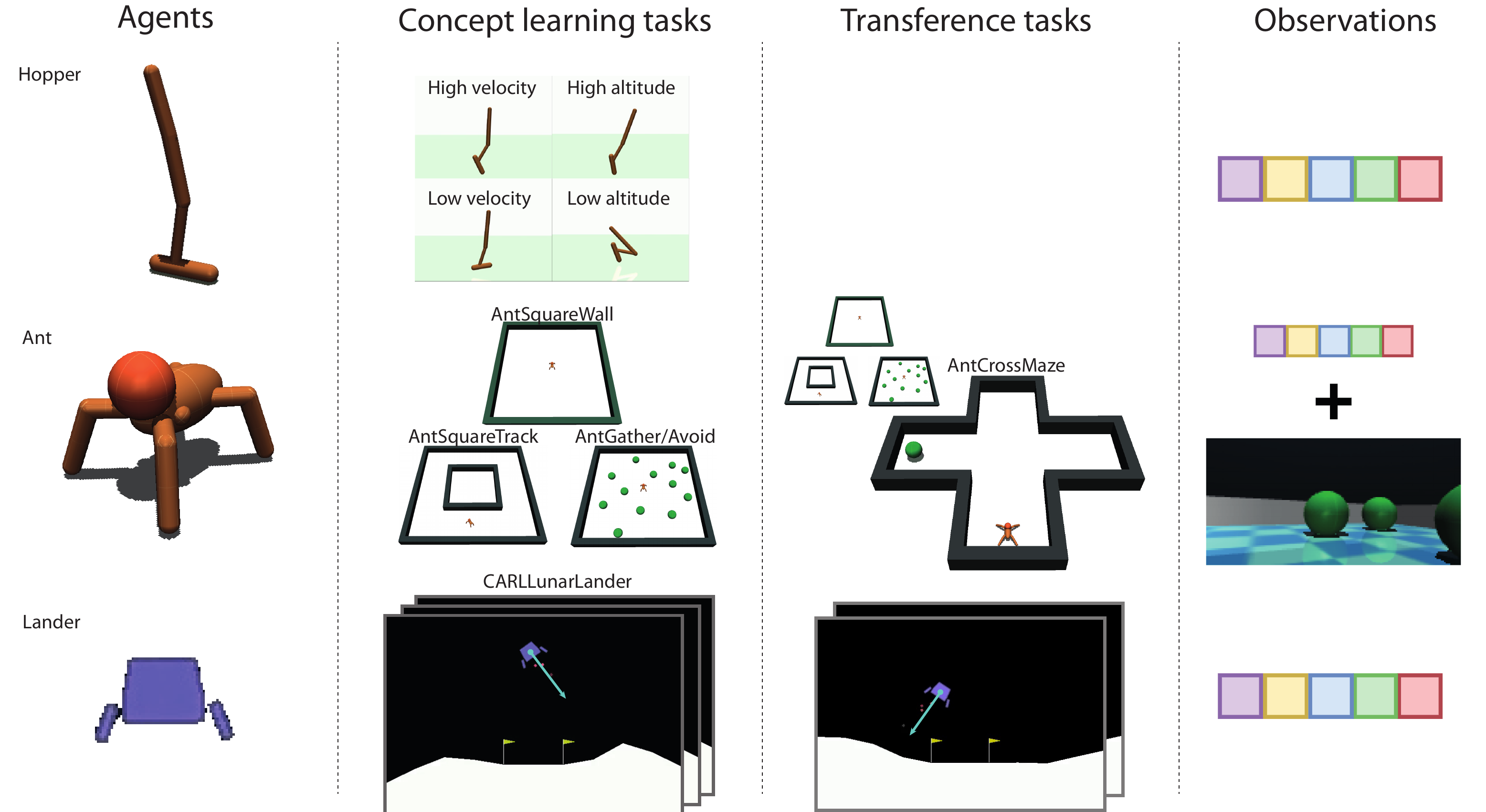}
\caption[]{Agents and tasks used to learn concepts. Concept learning tasks: Hopper: in all tasks the agent is rewarded by moving horizontally to the right. The reward is larger if the height or the velocity of the agent is high or low, depending on the task; Ant: in all of the tasks the ant is rewarded by moving as fast as possible. In the tasks with green spherical targets, the ant receives a positive or a negative reward when it is inside a target and these targets last for a finite amount of steps; Lander: the agent is rewarded by getting close to earth and landing satisfactorily, and it is punished by crashing and spending fuel. Transference tasks: Ant: the agent is re-trained in the previous tasks as well as in a new maze with cross shape. The ant only receives a positive reward when it reaches the exit, which is a green spherical target\footnotemark; Lander: the agent uses its concepts in two landing tasks with gravitational fields not previously experienced.}
\label{fig:tasks}
\end{figure*}


\subsection{Agents}

\textbf{Libraries:} We simulated locomotive environments with the MuJoCo physics engine \cite{todorov2012mujoco}, using the OpenAI gym wrapper \cite{brockman2016openai}. We made use of two agents: the Hopper agent from the OpenAI gym library and the Ant agent from \cite{frans2018meta}. We also used the CARL benchmark \cite{BenEim2021a} to generate several tasks with different dynamics for the environment \texttt{LunarLander} (also part of the OpenAI gym library). 

\textbf{Hopper:} The standard Hopper environment consists of a one-legged agent with three rotational actuators that is restricted to move in the $xz$-plane. Its objective is to hop towards the $+x$ direction. When the agent reaches certain height and angle thresholds it is restarted. The observations corresponds to a real vector with information on the position and velocities of the joints (Fig. \ref{fig:tasks}).  

\textbf{Ant:} The Ant agent consists of a semi-spherical body with 4 legs attached and a head. Its actuators correspond to 8 joints, 2 for each leg. The observations consisted of two parts: a real vector with information on the position and velocities of the joints, and a second element with visual local information. The visual data was represented as RGBA pixel images of size $84\times168\times3$ (Fig. \ref{fig:tasks}).

\textbf{Lander:} The Lander agent consists of a main body, two landing supports, a fire engine below the main body and two additional side engines. The objective is to land in a specific landing pad by firing the engines. The observations are a real vector with information about the position, velocity and contact sensors in the landing supports (Fig. \ref{fig:tasks}).

\subsection{Tasks and Techniques}
\textbf{Skill learning:} Since our proposed method initially only considers discrete actions and the action space is continuous in the locomotive environments considered, we learned a finite set of skills $\mathcal{Z}$ to replace the actions in the concept learning algorithms. Each skill $z\in \mathcal{Z}$ corresponds to a distribution of actions given the state and the skill $\pi(a|z,s)$.

\textbf{Hopper training:} The hopper agent was trained in 4 different tasks that resulted from modifying the reset thresholds. This allowed the agent to perform more diverse movements, such as crouching. We trained skills and high-level policies $\pi(z|s,c)$ in parallel, following a similar scheme as the one explained in \cite{frans2018meta}. We used the soft actor-critic algorithm (SAC) \cite{haarnoja2018soft} to train both skills and high-level policies. Since the skill space is discrete, we used a non-conventional version of SAC (see Supplementary Material). After a fixed number of steps, the task was randomly changed during an episode to ensure a diversity of states in the initiation sets for all tasks. This forced the agent to make marked transitions of behavior, which resulted in disentangled skills and concepts. Moreover, the exploration parameter $\alpha$ that appears in SAC was learned, as explained in \cite{haarnoja2018soft}.

\textbf{Ant training:} The ant training consisted of 3 stages. In the first one, 3 skills were trained in parallel using SAC \cite{haarnoja2018soft}: walking along a goal direction, rotating to the left, and rotating to the right. To achieve this, an appropriate reward signal related with linear or angular velocities was used. 

In the second stage, 4 different policies from states to skills $\pi(z|s,c)$ were learned using the SAC algorithm and noisy dueling deep Q networks \cite{wang2016dueling, fortunato2018noisy}. Each policy corresponded to a different task where the agent was presented with two types of objects: walls and spherical green targets. In all of these tasks the agent received a reward signal that encouraged having a large linear velocity, and, in the two tasks where targets appeared, it encouraged the agent to either seek or avoid the targets (see Figs. \ref{fig:tasks}, \ref{fig:ant_concepts}). 

The trajectories $\{\tau_i=(c_i,s_{0,i},a_{0,i},r_{1,i},s_{1,i},\cdots)\}$ of the agent in these 4 tasks were stored and later used to learn the concept classifier $\phi$. To benchmark the proposed metric in Eq. (\ref{eq:proposed_metric}), we selected two alternative metrics:
\begin{enumerate}
    \item max. likelihood metric \cite{asadi2020learning}: $\max_\phi\Pi_{i}\text{Pr}(\tau_i,\phi,\pi_\phi)$,
    \item context independent metric: $\min_{\phi}I(S:A|S_\phi)$.
\end{enumerate}
The classifier $\phi$ was learned by optimizing each of these metrics and subsequently the agent was trained in the same 4 tasks, but in this case high-level policies $\pi_\phi(z|s_\phi,c)$ were learned with the Trust Region Monte Carlo (TRMC) method introduced in Section \ref{sec:concepts}. To distinguish the different metrics we will refer to the learned concepts as \textit{information concepts} or \textit{likelihood concepts}. In addition, we used the Reptile first order meta-learning algorithm \cite{nichol2018first} to learn an appropriate weight initialization for the SAC algorithm. Since these weights allow the agent to transfer complete value estimates and policies, we performed an additional experiment with an ablation. We transferred only the weights from convolutional layers and we fixed these weights. In this way, we try to understand if the meta-learning approach is able to learn transferable state abstractions.


In the final stage, the Ant was trained to escape from a maze (see Fig. \ref{fig:tasks}), combining SAC with random network distillation (RND) \cite{burda2019exploration}. The second method is necessary in this case since the rewards are too sparse and we need some way to foster exploration. As in the previous case, the agent was trained with and without the use of the concepts. In this case, however, the abstract policy $\pi_\phi$ was used as a behavior prior to train the policy $\pi$ (see Supplementary Material).

\textbf{Lander training:} The \texttt{LunarLander} environment was solved for three different constant gravitational fields $(g_x,g_y)$ $\in$ $\{(-5,-6),(0,-10),(5,-6)\}$, by using dueling deep Q networks (DDQN) \cite{wang2016dueling}. As in the previous cases, the classifier $\phi$ was learned off-policy with the previously generated trajectories. Then, the concepts where used in two new landing tasks with $(g_x,g_y)$ $\in$ $\{(-4,-8),(4,-8)\}$.

\footnotetext{We replaced the green spheres by boxes, since MuJoCo makes transparent the spheres once an agent is inside.}

\subsection{Results}
\subsubsection{Qualitative concept assessment}
\begin{figure}
\centering
\includegraphics[width=0.93\columnwidth]{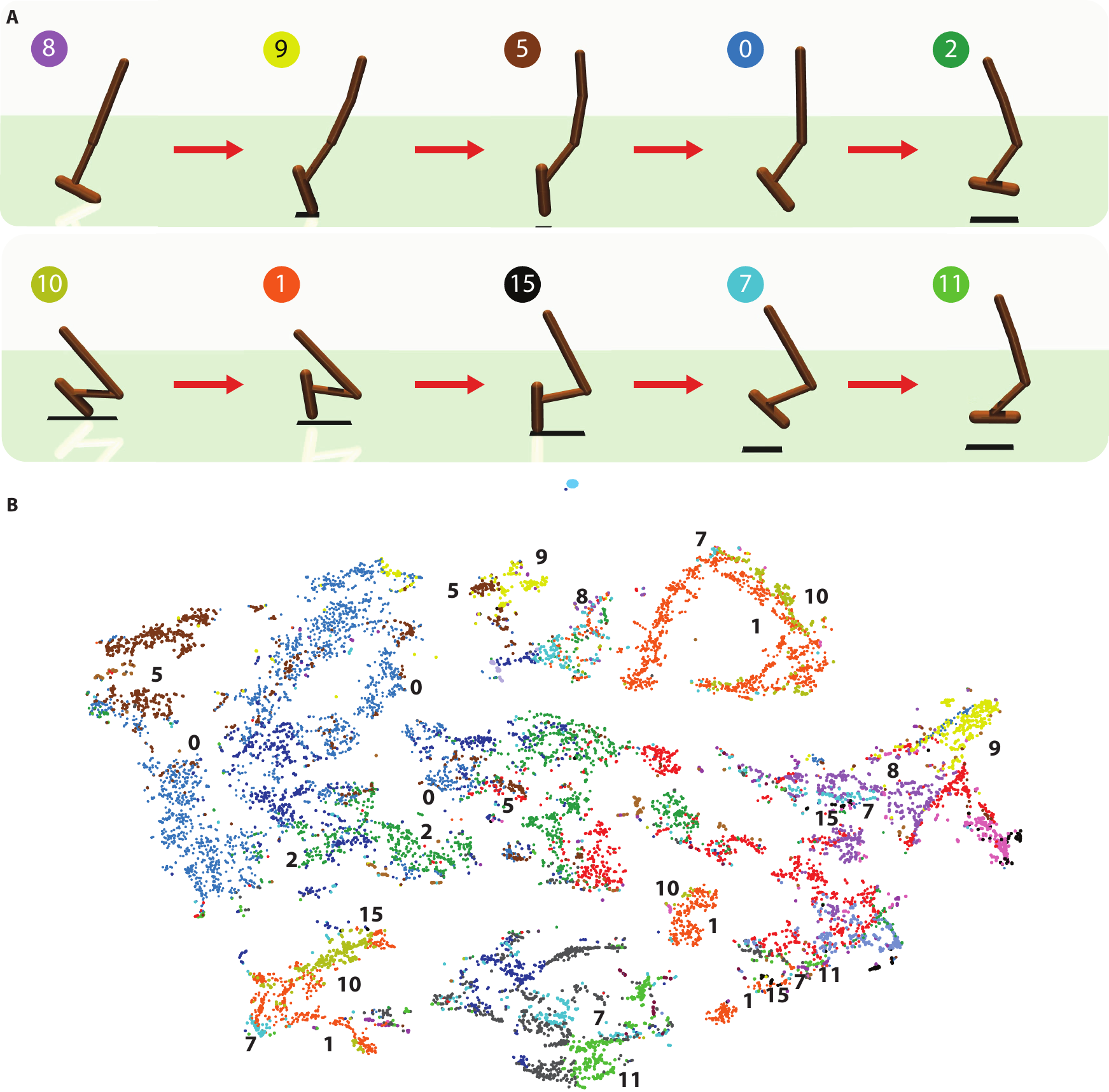}
\caption{Concepts learned by the Hopper agent after optimizing Eq. (\ref{eq:proposed_metric}). The agent is able to abstract concepts corresponding to shapes (\textit{e.g.}, leg extended, flexed, and inclined) and motions (\textit{e.g.}, jumping, landing, and falling) (A) Progression of concepts activated during two trajectories. The number indicates the concept most closely related to the state shown. (B) t-SNE projection of several trajectories. The color of each point corresponds to its associated concept, and the colors are the same as those used in (A). It can be noticed that concepts provide a disentangled partition of the states. This indicates that the policies $\pi(z|s,c)$ do provide a natural partition of the state-space. Now, while the states are being separated in regions, this separation does not seem to depend strongly on the distance of the original configurations. Furthermore, the projection of progressive steps in (A) appear close in multiple locations, not a single one. The fact that the similarity of these steps is identified for different configurations indicates that the concepts are not dividing the trajectories arbitrarily.} \label{fig:hopper_concepts}
\end{figure}

\begin{figure*}
\centering
\includegraphics[width=0.8\linewidth]{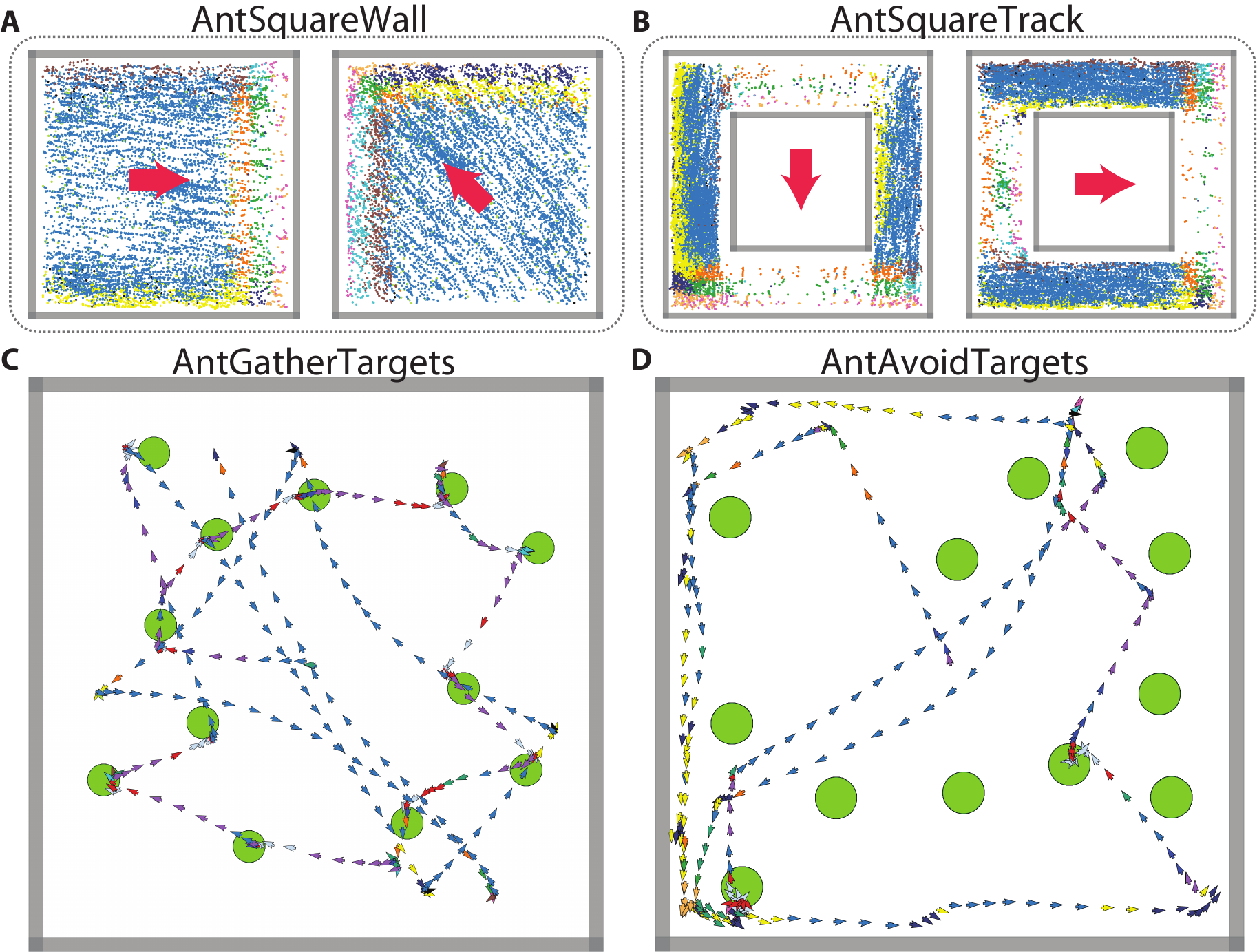}
\caption{Concepts learned by the Ant agent. In this case, the classifier is able to learn concepts like having a free path, being in front, to the right, or to the left of a wall, or being close or inside of a target. In all cases the points represent the $xy$ position of the ant and their color the concept identified. The arrows indicate the orientation of the ant. Several common concepts can be identified across the tasks: red - inside/in front of target, purple - target in the line of sight,  orange - wall in front (far), green - wall in front (close), blue - free path to walk, yellow - wall to the right, brown - wall to the left, \textit{etc}. In (A, B) the points correspond to samples stored in memory, while in (C, D) they are trajectories followed by the ant in a single episode. Additionally, in (C) the ant enters the targets since they correspond to a positive reward, while in (D) the opposite happens.} 
\label{fig:ant_concepts}
\end{figure*}

Figs. \ref{fig:hopper_concepts} and \ref{fig:ant_concepts} show the concepts learned by the Hopper and the Ant agents. We noticed that, in general, the learned concepts did not simply divide the state-space arbitrarily, but they managed to capture semantically meaningful configurations. This means that two states that could be distant in the Euclidean space but close in a semantic space would be effectively grouped together. This is supported by the t-SNE projection in Fig. \ref{fig:hopper_concepts}B, which shows distinct concepts that are non-trivially distributed along the state-space while maintaining a notion of distance between the concepts. In the case of the Hopper agent, the concepts learned are mainly related with shapes and motions of the body of the agent, while in the case of the Ant agent, they are related with external objects and their relative position to the agent. 

In Fig. \ref{fig:ant_concepts} we can observe that the agent manages to learn composite concepts like \lq wall to the right is moderately close\rq (points in yellow) or \lq target is in front\rq (points in red). Also, these concepts are consistent across different tasks, i.e., a discrete representation in one task, say the yellow one, corresponds to the same concept in another task. This suggests that the learned concepts are transferable and that they should indeed accelerate the learning of policies. Moreover, in the Supplementary Material it can be found a figure that presents the concepts learned when the random variable $S_\phi$ is separated in two variables $U_1$ and $U_2$, each with its corresponding classifier. The only additional information provided to the agent was the number of concepts for each classifier. The first variable seems to capture the idea of \textit{object} and is able to distinguish, in most cases, the walls, the targets and the absence of objects. In contrast, the second classifier seems to encode the idea of \textit{position} and is able to distinguish the relative position of the objects perceived, although it seems to fail more frequently than the first classifier. All in all, our results suggest that optimizing the objective in Eq. (\ref{eq:proposed_metric}) encourages learning compositional discrete representations that capture simple ideas and which can be combined to produce more complex concepts.

\subsubsection{Transference to known tasks}

\begin{figure*}
\centering
\includegraphics[width=0.87\linewidth]{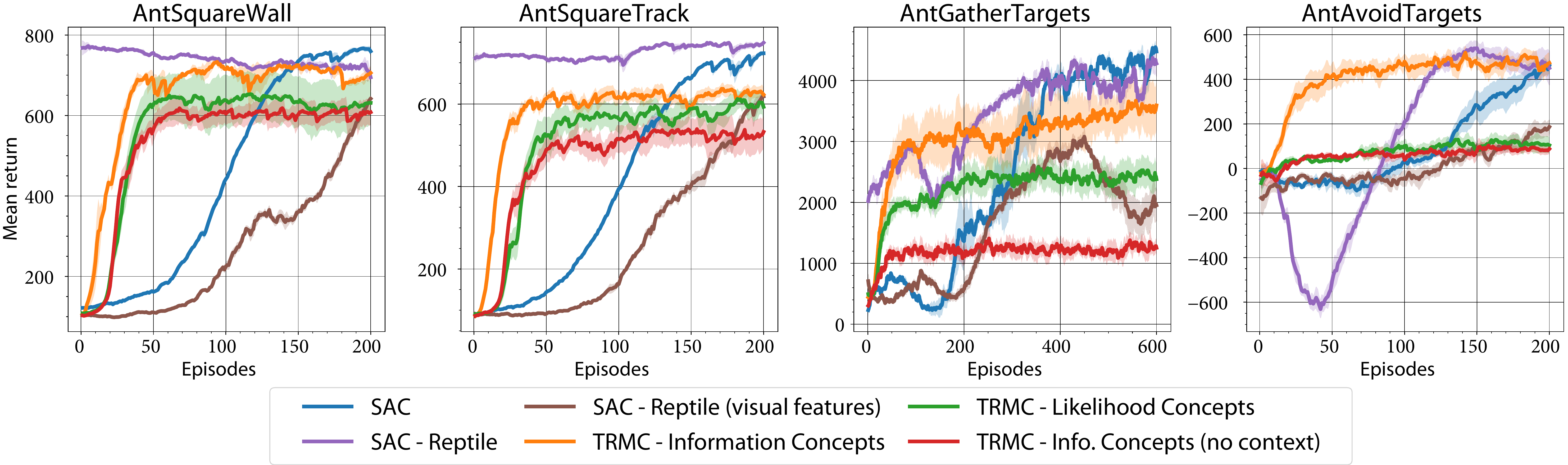}
\caption{Policy learning curves of the Ant agent in the 4 tasks where the concepts are learned. The blue, purple, and brown curves correspond to the case where the agent uses the baseline method SAC. The purple and brown lines include the Reptile weight initialization. The rest of the curves correspond to the cases where concepts are used to learn high-level policies $\pi_\phi$ through TRMC. Each curve corresponds to the average of 8 different seeds and the shades represent the population standard error.} 
\label{fig:comparison_multitask}
\end{figure*}

The learning curves of the Ant agent in the 4 tasks that were used to learn the concepts are shown in Fig. \ref{fig:comparison_multitask}. The performance $J_{\mathcal{M}_\text{test}}$ for the different algorithms can be measured in several ways \cite{taylor2009transfer, zhu2020transfer}. We will concentrate on three of them: the jumpstart $J_{\text{jp}}$, the asymptotic performance $J_{\text{ap}}$, and the time to threshold $J_{\text{tt}}$. The jumpstart corresponds to the initial return gap between the transfer learning curves and the baseline learning curve (SAC). We can see that $J_{\text{jp}}\approx0$ for all three concept metrics (TRMC). This is the case since the concepts do not store any information about the policy, even if the agent is learning in previously seen tasks. Similarly, the asymptotic performance refers to the final return gap with respect to the baseline. Table \ref{tab:perf_CP} shows $J_{\text{ap}}$ for all the tasks and learning algorithms. Being positive in most cases, it indicates that substituting the policy $\pi$ by the abstract policy $\pi_\phi$ results in a performance loss, ranging from $-4\%$ to $23\%$ for our proposed method. This is expectable since restraining the observations to the concepts results in throwing away some information which might be necessary for an optimal behavior. Nonetheless, the abstract policy $\pi_\phi$ can be used to accelerate the off-policy training of the policy $\pi$ and in the long run one could replace one for the other to avoid a negative $J_{\text{ap}}$ (Fig. \ref{fig:comparison}). The third metric, the time to threshold $J_{\text{tt}}$, makes explicit the trade-off between the loss of performance and the gain in efficiency. We define it as the ratio of learning steps or episodes that is required to achieve a particular performance threshold between the baseline and the other learning algorithms. We summarize these ratios in Table \ref{tab:n_eps} for two different thresholds. In all but one case, using the concepts learned by optimizing Eq. (\ref{eq:proposed_metric}) resulted in learning at least 3.9 times faster.

\begin{table}
\caption{Asymptotic performance $J_{ap}$ of the Ant agent after transferring its concepts or initialization weights. The numbers shown correspond to the difference of population means between the SAC baseline and each algorithm. The difference was averaged for the last 10 steps, to account for noise. The numbers in parenthesis are standard errors and the bold font indicates which method is statistically better.}
\label{tab:perf_CP}
\centering
\resizebox{\columnwidth}{!}{%
\begin{tabular}{|r|cccc|}
\hline
\multicolumn{1}{|c|}{\textbf{Task}} & \multicolumn{1}{c}{\textbf{AntSquareWall}} & \multicolumn{1}{c}{\textbf{AntSquareTrack}} & \multicolumn{1}{c}{\textbf{AntGather.}} & \textbf{AntAvoid.} \\ \hline \hline
TRMC - InfoConcepts         & \textbf{69.0} (11.1)  & 85.7 (13.0)  & 965.3 (305.0)  & \textbf{-19.8} (42.5) \\ 
TRMC - LikelihoodConcepts    & 137.2 (41.7) & 115.9 (23.7) & 2095.2 (189.2) & 332.0 (40.9) \\
TRMC-InfoConcepts (no context)    & 160.6 (24.1) & 190.2 (30.6) & 3250.5 (102.0) & 356.7 (33.0) \\ 
SAC-Reptile        & \textbf{48.0} (10.7)  & \textbf{-30.6} (7.8)  & \textbf{199.9} (129.6)  & \textbf{-26.2} (54.9) \\ 
SAC - Reptile (visual features) & 151.4 (8.0)  & 112.6 (12.8) & 2499.7 (199.6) & 267.2 (38.0) \\ \hline
\end{tabular}%
}
\end{table}


\begin{table}
\caption{Number of learning episodes for the Ant agent to reach $50\%$ and $80\%$ of the baseline maximum return. The ratio $J_{\text{tt}}$ between the baseline and the transfer number of episodes is shown in parenthesis.}
\label{tab:n_eps}
\centering
\resizebox{\columnwidth}{!}{%
\begin{tabular}{|r|cccc|}
\hline
\multicolumn{1}{|c|}{\textbf{Task}} & \multicolumn{1}{c}{\textbf{AntSquareWall}} & \multicolumn{1}{c}{\textbf{AntSquareTrack}} & \multicolumn{1}{c}{\textbf{AntGather.}} & \textbf{AntAvoid.} \\ \hline \hline
\multicolumn{1}{|c|}{\textbf{Episodes to reach 50\% of max.}} &  &  &  &  \\
SAC                       & 93           & 95           & 275         & 143 \\
TRMC - InfoConcepts              & 16 (5.8x)    & 14 (6.8x)    & 41 (6.7x)   & \textbf{19 (7.5x)} \\ 
TRMC - LikelihoodConcepts        & 28 (3.3x)    & 32 (3.0x)    & 156 (1.8x)  & Not reached \\ 
TRMC-InfoConcepts (no context) & 25 (3.7x)    & 26 (3.7x)    & Not reached & Not reached \\ 
SAC - Reptile              & \textbf{0 (}$\boldsymbol{\infty}$\textbf{)} & \textbf{0 (}$\boldsymbol{\infty}$\textbf{)} & \textbf{10 (27.5x)}  & 102 (1.4x) \\
SAC - Reptile (only features)   & 155 (0.6x)   & 133 (0.7x)   & 317 (0.9x)  & Not reached \\ \hline
\hline
\multicolumn{1}{|c|}{\textbf{Episodes to reach 80\% of max.}} &  &  &  &  \\
SAC                       & 122         & 129         & 363         & 180  \\
TRMC - InfoConcepts              & 31 (3.9x)   & 32 (4.0x)   & 541 (0.7x)  & \textbf{34 (5.3x)}  \\ 
TRMC - LikelihoodConcepts        & 45 (2.7x)   & 84 (1.5x)   & Not reached & Not reached \\ 
TRMC-InfoConcepts (no context) & 67 (1.8x)   & Not reached & Not reached & Not reached \\ 
SAC - Reptile              & \textbf{0 (}$\boldsymbol{\infty}$\textbf{)}    & \textbf{0 (}$\boldsymbol{\infty}$\textbf{)}    & \textbf{273 (1.3x)}  & 112 (1.6x)   \\
SAC - Reptile (visual features)   & 195 (0.6x)  & 187 (0.7x)  & Not reached & Not reached \\ \hline
\end{tabular}%
}
\end{table}

Considering the different performance results, we can conclude that our proposed metric results in learning concepts better suited for transfer learning than the other two state abstraction approaches. With respect to the maximum likelihood metric, maximizing how likely are the trajectories under the classifier $\phi$ and the abstract policy $\pi_\phi$ is in essence very similar to optimize the mutual information. The difference is that it is not obvious how to select $\pi_\phi$ in this case, while \hyperref[th1]{Theorem 1}  offers an analytical expression for $\pi_\phi^*$. In \cite{asadi2020learning} it is stated that the policy can be either optimized or assumed given. With no optimization method available, we opted to use a randomly generated policy $\pi_\phi$ and we can see that optimizing just the classifier $\phi$ is not enough to learn appropriate concepts. In particular, the fact that this approach failed to solve the \texttt{AntAvoidTargets} task indicates that the agent failed to learn concepts semantically associated with the green targets (Fig. \ref{fig:ant_concepts}). Something similar happens with the metric that optimizes the mutual information without taking into account the context. In this case, the agent fails in both tasks where the green targets appear. The problem is that by ignoring the context, it is implicitly assumed that for a specific state $s$ an agent should behave the same in all tasks in $\mathcal{C}$. This is clearly not the case here, given that the reward changes drastically between tasks and while in one task it is desirable to get close to the green targets, in the other one the opposite happens. This translates into learning interference that prevents learning concepts associated to the targets. For this reason, it is important to consider the concepts as task-equivariant sets of behavior, as opposed to only invariant sets. Something which is typically assumed when the task is not readily available to the agent. Moreover, the Reptile curves (purple) show that through first-order meta-learning the weights can remember the optimal policy for the two tasks where the green targets do not appear. On the contrary, this initialization seems unsuitable to handle tasks with contradictory policies. This can be deduced from the large negative return that the agent obtains in the first episodes of the \texttt{AntAvoidTargets} task, episodes during which it probably re-learns its value estimates and policy for observations related with the green targets. The second group of Reptile curves (brown), where the agent only transfers visual features, presents the worst behavior in almost all tasks. This allow us to conclude that transferable state representations are not being learned through meta-learning and that if the reward function or the dynamics change drastically between tasks this approach may fail, in contrast with our proposed method.

Now, it could be reasonably argued that what the concept networks are doing is actually generating a different partition for each task, and then they would be simply memorizing the optimal policies learned in the first training stage. This could easily happen if, for example, the states visited in the training tasks are not common. To guarantee that this is not what is happening, we estimated the entropy of the concepts $H(S_\phi)$ and the mutual information between concepts and tasks $I(S_\phi:C)$. If the concepts corresponded to disjoint partitions of the state-space, one for each task, then we would expect the mutual information $I(S_\phi:C)$ to be close to the concept entropy $H(S_\phi)$, given that knowing the task would largely reduce the uncertainty about the concept. In this case, the estimated entropy was $H(S_\phi)=2.44$, while the mutual information was $I(S_\phi:C)=0.19$. The large difference between these two quantities allows as to conclude that the concept network is not memorizing any information about the specific behavior of each task, but it is learning common representations among the tasks that facilitate the process of policy learning.

\subsubsection{Transference to unknown tasks}
\begin{figure}
\centering
\includegraphics[width=0.9\linewidth]{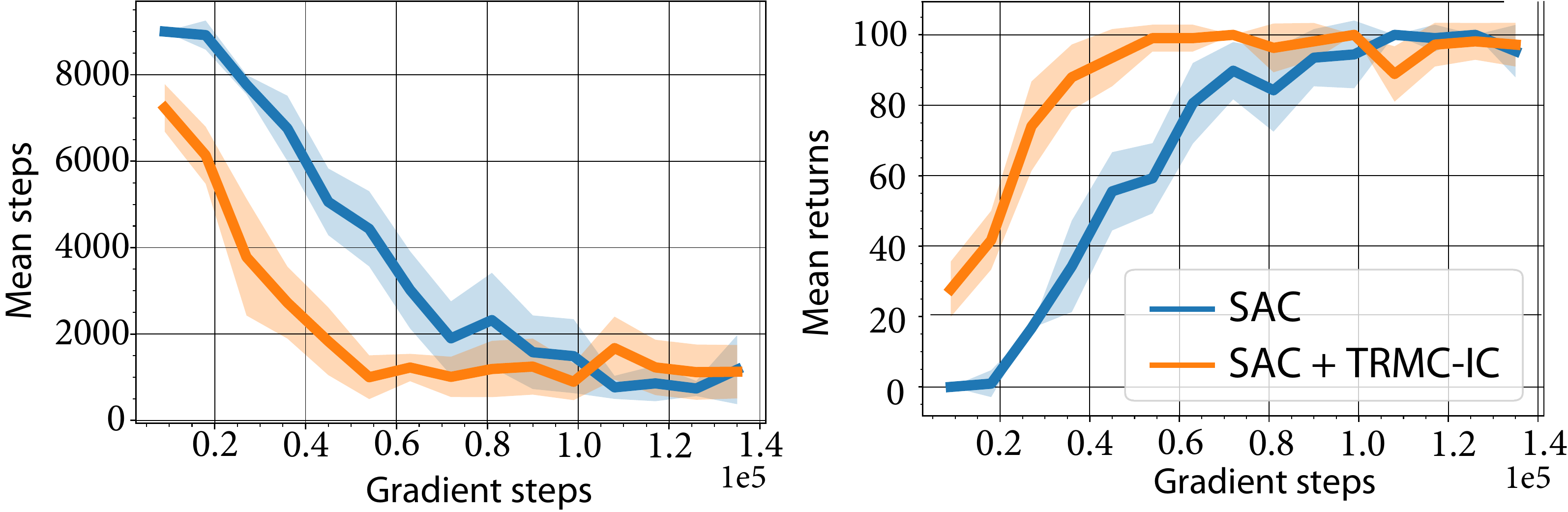} 
\caption{Concept transference to the \texttt{AntCrossMaze} task. We compared the performances of the standard method SAC and the same method endowed with information concepts (IC). Each curve corresponds to the average of 6 different seeds and 18 test episodes per data-point. The shades represent the population standard error. The number of steps required to escape from the maze decreases faster when the concepts are used as priors. Similarly, the mean returns increase faster since the agent is successful more times in finding the exit.}  
\label{fig:comparison}
\end{figure}

In Fig. \ref{fig:comparison} we can see the training curves of the Ant in a maze environment it had not seen previously. While the agent learned effectively in both cases, the convergence was 2 times faster (to reach $85\%$ of max. return) when the concepts were used. This result validates that our algorithm allows learning concepts that can be transferred to tasks not faced before and that present drastically different reward functions. This was expected in this case since all the objects appearing in the new task  (i.e., walls and targets) were also presented to the agent in previous tasks. Consequently, the bound $D(s_{\max},s'_{\max}|c_{\max})$  in \hyperref[cor1]{Corollary 1} should be small, and the fact that the state visitation is different did not affect radically the improvement in sample efficiency.

\begin{figure}
\centering
\includegraphics[width=0.81\linewidth]{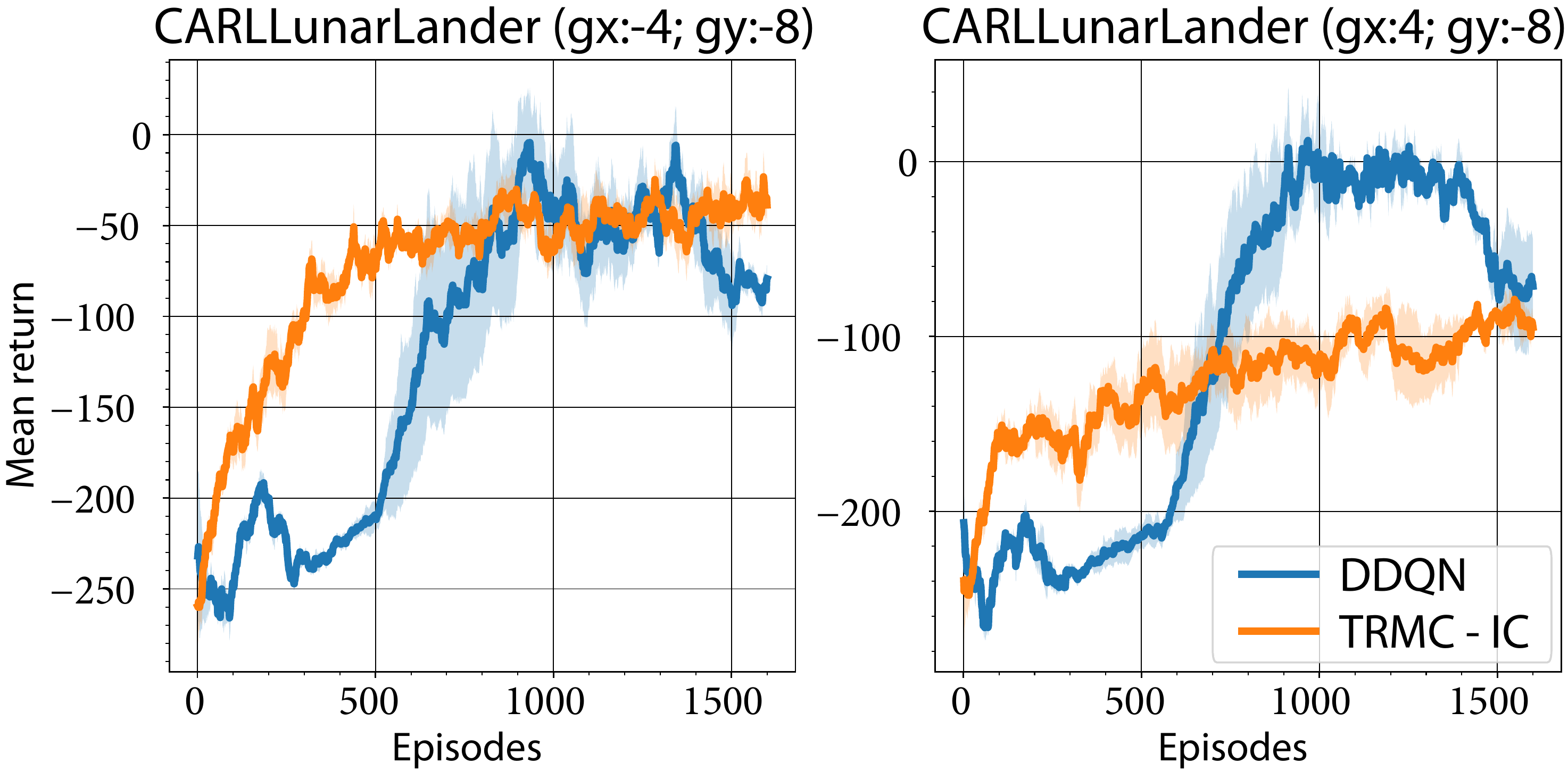}
\caption{Policy learning curves of the Lunar Lander agent in 2 unknown tasks. The titles indicate the components $g_x$ and $g_y$ of the gravitational field. The blue curves correspond to the case where the agent uses the baseline method DDQN. The orange ones correspond to the cases where information concepts (IC) are used to learn high-level policies $\pi_\phi$. Each curve corresponds to the average of 4 different seeds and the shades represent the population standard error.} 
\label{fig:comparison_lunar}
\end{figure}

In contrast, in the case of the Lander agent, Fig. \ref{fig:comparison_lunar} shows ambiguous results. In one of the tasks, the agent is markedly more sample efficient than the baseline, and even the final performance $J_{\text{ap}}$ is positive. On the other hand, for the other task the sample efficiency is only higher for a short period of time. The performance itself of the baseline was not satisfactory, and we observed that it acted randomly for some states. We hypothesise that using a suboptimal policy results in behavior that is not amenable to represent the state space through concepts, given that random behavior does not translate between tasks. In addition, something worth noting is that while the mutual information $I(Z:S|S_\phi,C)$ in the case of the Ant tasks was below $0.4$, in the case of the Lander agent, it was above $0.9$. This makes sense if one considers that the Ant tasks correspond directly to discrete semantic objectives such as ``avoid walls'' or ``touch targets''. 


\section{Discussion}
The first thing to point out of from the results is that the representations learned by using our algorithm are, to a certain extent, interpretable. While this could be expected by the simplicity of the tasks used, there is in principle no reason why this should be the case. After all, no constrain nor explicit bias is being used to obtain disentangled representations. The tasks used to learn the concepts were subject to the simulation of physical rules that determined the optimal behavior. Since the transferability of this optimal behavior was what guided the concept learning process, we believe our results are evidence that common tasks occupy a very particular region of the possible task space where the sensory inputs can be encoded with very simple representations, and in particular, discrete representations, that capture notions like equivariance and objectness. 

Our positive results showing that conceptual representations can be learned by solving control tasks is consistent with the neurological evidence that indicates that a good portion of the concepts that humans learn, \textit{concrete concepts}, are grounded in the sensory-motor brain systems \cite{kiefer2012Conceptual}. This type of concepts is closely related to the notions of object and action, like wall, walking, or distant, so it was expected that precisely the concepts of this type would be the ones learned. It remains to see if more abstract concepts can be learned in a similar fashion. A positive outcome would strengthen the embodied position that, while abstract concepts can be detached from any sensorimotor representation, they could in fact be learned by the same means than concrete concepts. Most possibly, learning this type of concepts requires considering further inductive biases concerning aspects like modularity or causal relations \cite{parascandolo2018learning}.

From a representation learning perspective, our algorithm can be seen as a self-supervised learning approach where the structures of the control problems faced by the agent are exploited to learn explainable discrete representations, just as other methods leverage the geometric structure of the environment \cite{kulkarni2019unsupervised}. In particular, our method can be readily combined with any skill-based hierarchical method or any standard method where the action space is discrete since it is an off-policy method. This property makes our method a useful tool to interpret what is being learned by current deep reinforcement learning methods, thus allowing to extract conceptual knowledge from them. 



With respect to the sample efficiency gained when using the knowledge of concepts in previously faced tasks, we consider that this supports an intermediate approach to the problem of multi-task and continual learning. Instead of trying to remember everything \cite{cully2015robots, heess2017emergence} or nothing, it could be more efficient remembering just the information that is common to most tasks and that facilitates discovering the optimal behavior. As seen in our results, this might result in a short term trade-off between sample efficiency and performance, but in principle low-level policies could be further refined in the long term if needed, in a similar way as how we humans gradually stop thinking about repetitive tasks. Moreover, our results are also in line with current studies that show that forgetting is not just a brain functioning error, but it is also the deliberate result of specific mechanisms like neurogenesis, which are necessary to form new memories and avoid interference with previous ones \cite{epp2016neurogenesis, moreno2020molecular}.


Finally, our algorithm could be considered as an inverse reinforcement learning algorithm where someone provides optimal trajectories and a model has to approximate this optimal behavior. In this sense, our work is very close to others like \cite{asadi2020learning, ajay2020opal}. Aside from introducing the concept of mutual information in this context, our work also sheds light in a common assumption when learning representations of the state-space and that is that the representations should determine the actions taken, irrespective of the context provided by the task. As shown in our experiments, this is overly restrictive and inhibits learning useful concepts like objects that accept multiple types of interactions. 


\section{Conclusions}
We introduced a principled deep reinforcement learning method that is able to learn conceptual representations of sensory inputs in a multi-task control setting, by minimizing a mutual information loss. We showed that these abstractions can capture composable intuitive concepts and improve the sample efficiency of standard algorithms, being at least twice as fast. Our experiments also validated that the concepts can be transferred to tasks with different reward functions and with different dynamics and still enjoy from a better sample efficiency. Based on our results, we consider that learning concepts could contribute notably to artificial agents being able to generalize as well as living beings do.

\ifCLASSOPTIONcompsoc
  \section*{Acknowledgments}
\else
  \section*{Acknowledgment}
\fi

We would like to thank Daniel Ochoa and Juan Pablo Martínez for valuable discussions and feedback.

\ifCLASSOPTIONcaptionsoff
  \newpage
\fi



%



\bibliographystyle{IEEEtran}
\bibliography{IEEEabrv,references}

\begin{thebibliography}{10}
\providecommand{\url}[1]{#1}
\csname url@samestyle\endcsname
\providecommand{\newblock}{\relax}
\providecommand{\bibinfo}[2]{#2}
\providecommand{\BIBentrySTDinterwordspacing}{\spaceskip=0pt\relax}
\providecommand{\BIBentryALTinterwordstretchfactor}{4}
\providecommand{\BIBentryALTinterwordspacing}{\spaceskip=\fontdimen2\font plus
\BIBentryALTinterwordstretchfactor\fontdimen3\font minus
  \fontdimen4\font\relax}
\providecommand{\BIBforeignlanguage}[2]{{%
\expandafter\ifx\csname l@#1\endcsname\relax
\typeout{** WARNING: IEEEtran.bst: No hyphenation pattern has been}%
\typeout{** loaded for the language `#1'. Using the pattern for}%
\typeout{** the default language instead.}%
\else
\language=\csname l@#1\endcsname
\fi
#2}}
\providecommand{\BIBdecl}{\relax}
\BIBdecl

\bibitem{mery2004operative}
F.~Mery and T.~J. Kawecki, ``{An operating cost of learning in Drosophila
  melanogaster},'' \emph{Animal Behaviour}, vol.~68, pp. 589--598, 2004.

\bibitem{niven2016neuronal}
J.~E. Niven, ``{Neuronal energy consumption: biophysics, efficiency and
  evolution},'' \emph{Current Opinion in Neurobiology}, vol.~41, pp. 129--135,
  2016.

\bibitem{bongard2006resilient}
J.~Bongard \emph{et~al.}, ``Resilient machines through continuous
  self-modeling,'' \emph{Science}, vol. 314, no. 5802, pp. 1118--1121, 2006.

\bibitem{haarnoja2018soft}
T.~Haarnoja \emph{et~al.}, ``{Soft Actor-Critic Algorithms and Applications},''
  arXiv:1812.05905, 2018.

\bibitem{mnih2015human}
V.~Mnih \emph{et~al.}, ``{Human-level control through deep reinforcement
  learning},'' \emph{Nature}, vol. 518, no. 7540, p. 529, 2015.

\bibitem{silver2017mastering}
D.~Silver \emph{et~al.}, ``{Mastering the game of Go without human
  knowledge},'' \emph{Nature}, vol. 550, no. 7676, p. 354, 2017.

\bibitem{andrychowicz2020learning}
O.~M. Andrychowicz \emph{et~al.}, ``Learning dexterous in-hand manipulation,''
  \emph{The International Journal of Robotics Research}, vol.~39, no.~1, pp.
  3--20, 2020.

\bibitem{cully2015robots}
A.~Cully \emph{et~al.}, ``Robots that can adapt like animals,'' \emph{Nature},
  vol. 521, no. 7553, pp. 503--507, 2015.

\bibitem{flesch2018comparing}
T.~Flesch \emph{et~al.}, ``Comparing continual task learning in minds and
  machines,'' \emph{Proceedings of the National Academy of Sciences of the
  United States of America}, vol. 115, no.~44, pp. E10\,313--E10\,322, 2018.

\bibitem{cobbe2019quantifying}
K.~Cobbe \emph{et~al.}, ``{Quantifying Generalization in Reinforcement
  Learning},'' in \emph{Proceedings of the 36th ICML}, vol.~97, 2019, pp.
  1282--1289.

\bibitem{zhao2019investigating}
C.~Zhao \emph{et~al.}, ``{Investigating Generalisation in Continuous Deep
  Reinforcement Learning},'' arXiv:1902.07015, 2019.

\bibitem{harb2018waiting}
J.~Harb \emph{et~al.}, ``{When Waiting Is Not an Option: Learning Options With
  a Deliberation Cost},'' in \emph{Proceedings of the Thirty-Second {AAAI}
  Conference on Artificial Intelligence}, 2018, pp. 3165--3172.

\bibitem{duan2016benchmarking}
Y.~Duan \emph{et~al.}, ``{Benchmarking deep reinforcement learning for
  continuous control},'' in \emph{Proceedings of the 33rd ICML}, 2016, pp.
  1329--1338.

\bibitem{geirhos2020shortcut}
R.~Geirhos \emph{et~al.}, ``{Shortcut learning in deep neural networks},''
  \emph{Nature Machine Intelligence}, vol.~2, pp. 665--673, 2020.

\bibitem{zhang2008discrete}
W.~Zhang and S.~J. Luck, ``{Discrete fixed-resolution representations in visual
  working memory},'' \emph{Nature}, vol. 453, pp. 233--235, 2008.

\bibitem{linderman2019hierarchical}
S.~Linderman \emph{et~al.}, ``{Hierarchical recurrent state space models reveal
  discrete and continuous dynamics of neural activity in C. elegans},''
  bioRxiv:10.1101/621540, 2019.

\bibitem{higgins2020unsupervised}
I.~Higgins \emph{et~al.}, ``{Unsupervised deep learning identifies semantic
  disentanglement in single inferotemporal neurons},'' arXiv:2006.14304, 2019.

\bibitem{zaslavsky2018efficient}
N.~Zaslavsky \emph{et~al.}, ``{Efficient human-like semantic representations
  via the Information Bottleneck principle},'' arXiv:1808.03353, 2018.

\bibitem{briscoe2001evolution}
A.~D. Briscoe and L.~Chittka, ``{The evolution of color vision in insects.}''
  \emph{Annual Review of Entomology}, vol.~46, pp. 471--510, 2001.

\bibitem{hauser2002faculty}
M.~D. Hauser \emph{et~al.}, ``{The Faculty of Language: What Is It, Who Has It,
  and How Did It Evolve?}'' \emph{Science}, vol. 298, no. 5598, pp. 1569--1579,
  2002.

\bibitem{shanahan2020artificial}
M.~Shanahan \emph{et~al.}, ``Artificial intelligence and the common sense of
  animals,'' \emph{Trends in Cognitive Sciences}, 2020.

\bibitem{davidson2020investigating}
G.~Davidson and B.~M. Lake, ``{Investigating Simple Object Representations in
  Model-Free Deep Reinforcement Learning},'' arXiv:2002.06703, 2020.

\bibitem{tesauro2002programming}
G.~Tesauro, ``{Programming backgammon using self-teaching neural nets},''
  \emph{Artificial Intelligence}, vol. 134, no. 1-2, pp. 181--199, 2002.

\bibitem{hessel2019inductive}
M.~Hessel \emph{et~al.}, ``{On inductive biases in deep reinforcement
  learning},'' arXiv:1907.02908, 2019.

\bibitem{veerapaneni2019entity}
R.~Veerapaneni \emph{et~al.}, ``{Entity Abstraction in Visual Model-Based
  Reinforcement Learning},'' arXiv:1910.12827, 2019.

\bibitem{kulkarni2019unsupervised}
T.~D. Kulkarni \emph{et~al.}, ``{Unsupervised Learning of Object Keypoints for
  Perception and Control},'' in \emph{Advances in NeurIPS}, vol.~32, 2019, pp.
  10\,724--10\,734.

\bibitem{razavi2019generating}
A.~Razavi \emph{et~al.}, ``{Generating Diverse High-Fidelity Images with
  VQ-VAE-2},'' in \emph{Advances in NeurIPS}, vol.~32, 2019, pp.
  14\,866--14\,876.

\bibitem{sutton1999between}
R.~S. Sutton \emph{et~al.}, ``Between mdps and semi-mdps: A framework for
  temporal abstraction in reinforcement learning,'' \emph{Artificial
  intelligence}, vol. 112, no. 1-2, pp. 181--211, 1999.

\bibitem{bacon2017opctioncritic}
P.-L. Bacon \emph{et~al.}, ``{The Option-Critic Architecture},'' in
  \emph{Proceedings of the Thirty-First AAAI Conference on Artificial
  Intelligence}, 2017, p. 1726–1734.

\bibitem{frans2018meta}
K.~Frans \emph{et~al.}, ``{Meta Learning Shared Hierarchies},'' in \emph{ICLR},
  2018.

\bibitem{goyal2020reinforcement}
A.~Goyal \emph{et~al.}, ``Reinforcement learning with competitive ensembles of
  information-constrained primitives,'' in \emph{ICLR}, 2020.

\bibitem{hafner2020mastering}
D.~Hafner \emph{et~al.}, ``{Mastering atari with discrete world models},''
  arXiv:2010.02193, 2020.

\bibitem{kirk2021survey}
R.~Kirk \emph{et~al.}, ``{A survey of generalisation in deep reinforcement
  learning},'' arXiv:2111.09794, 2021.

\bibitem{taylor2009transfer}
M.~E. Taylor and P.~Stone, ``{Transfer Learning for Reinforcement Learning
  Domains: A survey},'' \emph{JMLR}, vol.~10, no.~7, 2009.

\bibitem{zhu2020transfer}
Z.~Zhu \emph{et~al.}, ``{Transfer Learning in Deep Reinforcement Learning: A
  survey},'' arXiv:2009.07888, 2020.

\bibitem{zhang2019deep}
M.~Zhang \emph{et~al.}, ``Deep latent low-rank representation for face sketch
  synthesis,'' \emph{IEEE Transactions on Neural Networks and Learning
  Systems}, vol.~30, no.~10, pp. 3109--3123, 2019.

\bibitem{yu2020gradient}
T.~Yu \emph{et~al.}, ``{Gradient Surgery for Multi-Task Learning},'' in
  \emph{Advances in NeurIPS}, vol.~33, 2020, pp. 5824--5836.

\bibitem{finn2017model}
C.~Finn \emph{et~al.}, ``{Model-Agnostic Meta-Learning for Fast Adaptation of
  Deep Networks},'' in \emph{Proceedings of the 34th ICML}, vol.~70, 2017, pp.
  1126--1135.

\bibitem{nichol2018first}
A.~Nichol \emph{et~al.}, ``{On first-order meta-learning algorithms},''
  arXiv:1803.02999, 2018.

\bibitem{kostrikov2019imitation}
I.~Kostrikov \emph{et~al.}, ``{Imitation Learning via Off-Policy Distribution
  Matching},'' arXiv:1912.05032, 2019.

\bibitem{riemer2018learning}
M.~Riemer \emph{et~al.}, ``{Learning Abstract Options},'' in \emph{Advances in
  NeurIPS}, vol.~31, 2018, pp. 10\,445--–10\,455.

\bibitem{eysenbach2019diversity}
B.~Eysenbach \emph{et~al.}, ``{Diversity is All You Need: Learning Skills
  without a Reward Function},'' in \emph{ICLR}, 2019.

\bibitem{singh1995reinforcement}
S.~P. Singh \emph{et~al.}, ``{Reinforcement Learning with Soft State
  Aggregation},'' in \emph{Advances in NeurIPS}, vol.~7, 1994, pp. 361--368.

\bibitem{li2006towards}
L.~Li \emph{et~al.}, ``{Towards a Unified Theory of State Abstraction for
  MDPs},'' \emph{ISAIM}, vol.~4, p.~5, 2006.

\bibitem{jong2005state}
N.~K. Jong and P.~Stone, ``State abstraction discovery from irrelevant state
  variables.'' in \emph{IJCAI}, vol.~8, 2005, pp. 752--757.

\bibitem{abel2016near}
D.~Abel \emph{et~al.}, ``{Near optimal behavior via approximate state
  abstraction},'' in \emph{Proceedings of the 33th ICML}, vol.~48, 2016, pp.
  2915--2923.

\bibitem{zhang2021learning}
A.~Zhang \emph{et~al.}, ``{Learning Invariant Representations for Reinforcement
  Learning without Reconstruction},'' in \emph{ICLR}, 2021.

\bibitem{baillargeon2008innate}
R.~Baillargeon, ``{Innate ideas revisited: For a principle of persistence in
  infants' physical reasoning},'' \emph{Perspectives on Psychological Science},
  vol.~3, no.~1, pp. 2--13, 2008.

\bibitem{fields2017eigenforms}
C.~Fields \emph{et~al.}, ``{Eigenforms, Interfaces and Holographic Encoding:
  Toward an Evolutionary Account of Objects and Spacetime},''
  \emph{Constructivist Foundations}, vol.~12, no.~3, pp. 265--274, 2017.

\bibitem{kingma2014auto}
D.~P. Kingma and M.~Welling, ``Auto-encoding variational bayes,'' in
  \emph{ICLR}, 2014.

\bibitem{linsker1988self}
R.~Linsker, ``Self-organization in a perceptual network,'' \emph{Computer},
  vol.~21, no.~3, pp. 105--117, 1988.

\bibitem{schmidhuber1992learning}
J.~Schmidhuber, ``Learning factorial codes by predictability minimization,''
  \emph{Neural computation}, vol.~4, no.~6, pp. 863--879, 1992.

\bibitem{tishby2000information}
N.~Tishby \emph{et~al.}, ``{The information bottleneck method},''
  arXiv:physics/0004057, 2000.

\bibitem{alemi2017deep}
A.~A. Alemi \emph{et~al.}, ``{Deep Variational Information Bottleneck},'' in
  \emph{ICLR}, 2017.

\bibitem{hafner2020action}
D.~Hafner \emph{et~al.}, ``{Action and perception as divergence
  minimization},'' arXiv:2009.01791, 2020.

\bibitem{goyal2019infobot}
A.~Goyal \emph{et~al.}, ``Infobot: Transfer and exploration via the information
  bottleneck,'' in \emph{ICLR}, 2019.

\bibitem{igl2019generalization}
M.~Igl \emph{et~al.}, ``{Generalization in Reinforcement Learning with
  Selective Noise Injection and Information Bottleneck},'' in \emph{Advances in
  NeurIPS}, vol.~32, 2019, pp. 13\,978--13\,990.

\bibitem{tishby2011information}
N.~Tishby and D.~Polani, ``Information theory of decisions and actions,'' in
  \emph{Perception-action cycle}.\hskip 1em plus 0.5em minus 0.4em\relax
  Springer, 2011, pp. 601--636.

\bibitem{rubin2012trading}
J.~Rubin \emph{et~al.}, ``Trading value and information in mdps,'' in
  \emph{Decision Making with Imperfect Decision Makers}.\hskip 1em plus 0.5em
  minus 0.4em\relax Springer, 2012, pp. 57--74.

\bibitem{abel2019state}
D.~Abel \emph{et~al.}, ``State abstraction as compression in apprenticeship
  learning,'' in \emph{Proceedings of the Thirty-Third AAAI Conference on
  Artificial Intelligence}, vol.~33, no.~01, 2019, pp. 3134--3142.

\bibitem{asadi2020learning}
K.~Asadi \emph{et~al.}, ``{Learning state abstractions for transfer in
  continuous control},'' arXiv:2002.05518, 2020.

\bibitem{sutton2018reinforcement}
R.~S. Sutton and A.~G. Barto, \emph{{Reinforcement Learning: An
  Introduction}}.\hskip 1em plus 0.5em minus 0.4em\relax Cambridge: MIT press,
  2018.

\bibitem{schulman2015trust}
J.~Schulman \emph{et~al.}, ``{Trust region policy optimization},'' in
  \emph{Proceedings of the 32nd ICML}, 2015, p. 1889–1897.

\bibitem{todorov2012mujoco}
E.~Todorov \emph{et~al.}, ``{MuJoCo: A physics engine for model-based
  control},'' \emph{Proceedings of the 2012 IEEE/RSJ International Conference
  on Intelligent Robots and Systems}, pp. 5026--5033, 2012.

\bibitem{brockman2016openai}
G.~Brockman \emph{et~al.}, ``{OpenAI Gym},'' arXiv:1606.01540, 2016.

\bibitem{BenEim2021a}
C.~Benjamins \emph{et~al.}, ``{CARL: A Benchmark for Contextual and Adaptive
  Reinforcement Learning},'' in \emph{NeurIPS 2021 Workshop on Ecological
  Theory of Reinforcement Learning}, Dec. 2021.

\bibitem{wang2016dueling}
Z.~Wang \emph{et~al.}, ``{Dueling Network Architectures for Deep Reinforcement
  Learning},'' in \emph{Proceedings of the 33rd ICML}, 2016, pp. 1995--2003.

\bibitem{fortunato2018noisy}
M.~Fortunato \emph{et~al.}, ``{Noisy Networks For Exploration},'' in
  \emph{ICLR}, 2018.

\bibitem{burda2019exploration}
Y.~Burda \emph{et~al.}, ``Exploration by random network distillation,'' in
  \emph{ICLR}, 2019.

\bibitem{kiefer2012Conceptual}
M.~Kiefer and F.~Pulverm{\"u}ller, ``{Conceptual representations in mind and
  brain: Theoretical developments, current evidence and future directions},''
  \emph{Cortex}, vol.~48, pp. 805--825, 2012.

\bibitem{parascandolo2018learning}
G.~Parascandolo \emph{et~al.}, ``{Learning Independent Causal Mechanisms},'' in
  \emph{Proceedings of the 35th ICML}, vol.~80, 2018, pp. 4033--4041.

\bibitem{heess2017emergence}
N.~Heess \emph{et~al.}, ``{Emergence of locomotion behaviours in rich
  environments},'' arXiv:1707.02286, 2017.

\bibitem{epp2016neurogenesis}
J.~R. Epp \emph{et~al.}, ``Neurogenesis-mediated forgetting minimizes proactive
  interference,'' \emph{Nature communications}, vol.~7, no.~1, pp. 1--8, 2016.

\bibitem{moreno2020molecular}
A.~Moreno, ``Molecular mechanisms of forgetting,'' \emph{European Journal of
  Neuroscience}, 2020.

\bibitem{ajay2020opal}
A.~Ajay \emph{et~al.}, ``{OPAL: Offline Primitive Discovery for Accelerating
  Offline Reinforcement Learning},'' arXiv:2010.13611, 2020.

\end{thebibliography}

%

\begin{IEEEbiography}[{\includegraphics[width=1in,height=1in,clip,keepaspectratio]{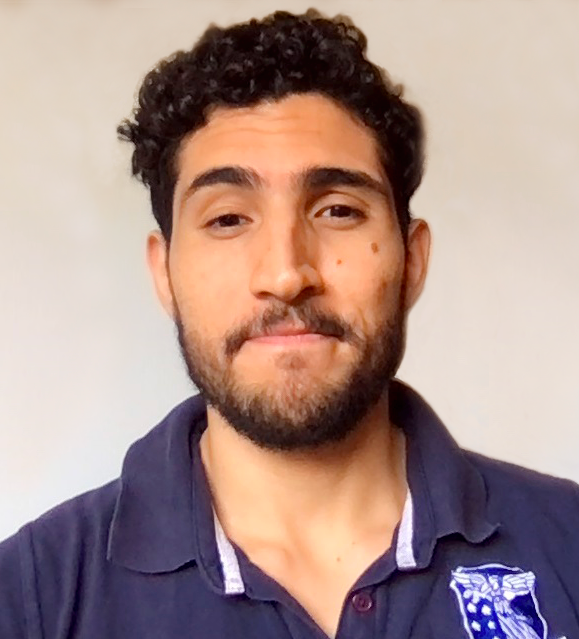}}]{Diego Gomez}
Diego Gomez received his B.Sc. degrees in Electronic Engineering and Physics from Universidad de Los Andes, Bogotá, Colombia, in 2019. He is currently a masters student at the same university. His current research interests include reinforcement learning, meta-learning, causal inference, and computer vision.
\end{IEEEbiography}


\begin{IEEEbiography}[{\includegraphics[width=1in,height=1.2in,clip,keepaspectratio]{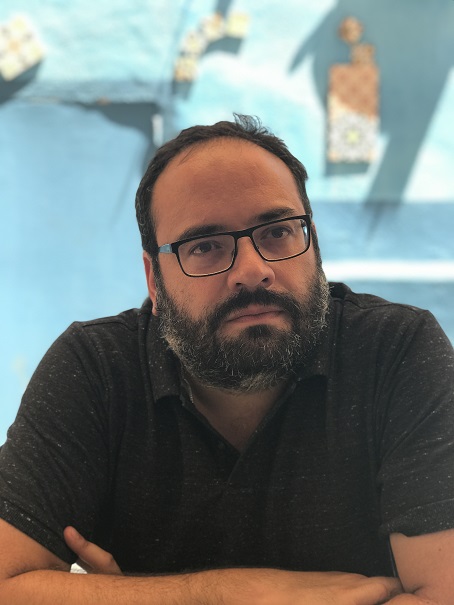}}]{Nicanor Quijano}
Nicanor Quijano (IEEE Senior Member) received his B.S. degree in Electronics Engineering from Pontificia Universidad Javeriana (PUJ), Bogotá, Colombia, in 1999. He received the M.S. and PhD degrees in Electrical and Computer Engineering from The Ohio State University, in 2002 and 2006, respectively. In 2007, he joined the Electrical and Electronics Engineering Department, Universidad de los Andes (UAndes), Bogotá, Colombia. He is currently a Full Professor, the director of the research group in control and automation systems (GIAP, UAndes), and an associate editor for the IEEE Transactions on Control Systems Technology, the Journal of Modern Power Systems and Clean Energy, and Energy Systems. On the other hand, he has been a member of the Board of Governors of the IEEE Control Systems Society (CSS) for the 2014 period, and he was the chair of the IEEE CSS, Colombia for the 2011-2013 period. He has published more than 30 journal papers, 70 international conference papers, and 5 book chapters. He has co-advised the best European PhD thesis in the control systems area in 2017, and he is the co-author of the best paper of the ISA Transactions, 2018. Currently his research interests include: hierarchical and distributed optimization methods using bio-inspired and game-theoretical techniques for dynamic resource allocation problems, especially those in energy, water, agriculture, and transportation. 

\end{IEEEbiography}


\begin{IEEEbiography}[{\includegraphics[width=1in,height=1in,clip,keepaspectratio]{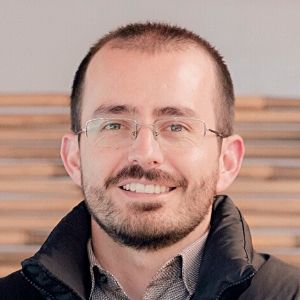}}]{Luis Felipe Giraldo}
Luis Felipe Giraldo (S’08, M’16) received the Ph.D. degree in electrical and computer engineering from The Ohio State University, Columbus, OH, USA, in 2016. 

He is an Associate Professor with the Universidad de los Andes, Bogotá, Colombia, and currently he is member of the Center for Artificial Intelligence Research and Education CINFONIA. His current research interests include design and analysis of interconnected dynamical systems and machine learning. 
\end{IEEEbiography}




\end{document}